\documentclass[10pt,journal,compsoc]{IEEEtran}
\usepackage{color}
\usepackage{times}

\newcommand{\eat}[1]{}
\usepackage[table,xcdraw]{xcolor}
\usepackage{times}
\usepackage{epsfig}
\usepackage{graphicx}
\usepackage{epstopdf}

\usepackage{amsmath}
\usepackage{amssymb}
\usepackage{rotating}
\usepackage{booktabs}
\usepackage{array}
\usepackage{booktabs}
\usepackage{color}
\usepackage{hyperref}

\usepackage{longtable}
\ifCLASSOPTIONcompsoc
  \usepackage[nocompress]{cite}
\else
  \usepackage{cite}
\fi

\begin{document}

\title{Hierarchical LSTMs with Adaptive Attention for Visual Captioning}

\author{Jingkuan Song,~\IEEEmembership{Member,~IEEE,}
        Xiangpeng Li,~\IEEEmembership{}
         Lianli Gao,~\IEEEmembership{}
        and Heng Tao Shen% <-this % stops a space
\IEEEcompsocitemizethanks{\IEEEcompsocthanksitem J. Song, X. Li, L. Gao and H. Shen are with the Future Media Center and School of Computer Science and Engineering, The University of Electronic Science and Technology of China, Chengdu, China, 611731. E-mail: lianli.gao@uestc.edu.cn
}
\thanks{Manuscript received April 19, 2005; revised August 26, 2015.}}

\markboth{Journal of \LaTeX\ Class Files,~Vol.~14, No.~8, August~2015}%
{Shell \MakeLowercase{\textit{et al.}}: Bare Demo of IEEEtran.cls for Computer Society Journals}

\IEEEtitleabstractindextext{
\begin{abstract}
Recent progress has been made in using attention based encoder-decoder framework for image and video captioning. Most existing decoders apply the attention mechanism to every generated word including both visual words (e.g., ``gun" and ``shooting") and non-visual words (e.g. ``the", ``a"). However, these non-visual words can be easily predicted using natural language model without considering visual signals or attention. Imposing attention mechanism on non-visual words could mislead and decrease the overall performance of visual captioning. 
Furthermore, the hierarchy of LSTMs enables more complex representation of visual data, capturing information at different scales.
To address these issues, we propose a hierarchical LSTM with adaptive attention (hLSTMat) approach for image and video captioning. Specifically, the proposed framework utilizes the spatial or temporal attention for selecting specific regions or frames to predict the related words, while the adaptive attention is for deciding whether to depend on the visual information or the language context information. 
Also, a hierarchical LSTMs is designed to simultaneously consider both low-level visual information and high-level language context information to support the caption generation. 
We initially design our hLSTMat for video captioning task. Then, we further refine it and apply it to image captioning task.  
% In addition, different features can provide complementary information to describe objects or actions, while each type of feature suffers from some limitations to handle the variations of objects or actions.  To further improve the video captioning performance, we extend hLSTMat to capture the complementary information on appearance from still frames and motion between frames through constructing variants of spatial-temporal hLSTMat networks, which fuse both spatially and temporally in order to best take advantage of the spatio-temporal information. 
To demonstrate the effectiveness of our proposed framework, we test our method on both video and image captioning tasks. Experimental results show that our approach achieves the state-of-the-art performance for most of the evaluation metrics on both tasks. The effect of important components is also well exploited in the ablation study.
\end{abstract}

}

\maketitle

\IEEEdisplaynontitleabstractindextext

\IEEEpeerreviewmaketitle

\IEEEraisesectionheading{\section{Introduction}\label{sec:introduction}}

\IEEEPARstart{P}{reviously}, visual content understanding ~\cite{song2016optimized,GaoWSHSS17} and natural language processing (NLP) are not correlative with each other. Integrating visual content with natural language learning to generate descriptions for images, especially for videos, has been regarded as a challenging task. Video captioning is a critical step towards machine intelligence and many applications in daily scenarios, such as video retrieval \cite{wang2017survey,song2017quantization}, video understanding, blind navigation and automatic video subtitling.  

\eat{\textcolor{red}{Why use LSTM}}

Thanks to the rapid development of deep Convolutional Neural Network (CNN), recent works have made significant progress for image captioning \cite{vinyals2015show,xu2015show,Lu2016Knowing,karpathy2014deep,fang2015captions,chen2014learning,chen2016sca}. However, compared with image captioning, video captioning is more difficult due to the diverse sets of objects, scenes, actions, attributes and salient contents. Despite the difficulty there have been a few attempts for video description generation \cite{venugopalan2014translating,venugopalan2015sequence,yao2015describing,li2015summarization,45493}, which are mainly inspired by recent advances in translating with Long Short-Term Memory (LSTM). The LSTM is proposed to overcome the vanishing gradients problem by enabling the network to learn when to forget previous hidden states and when to update hidden states by integrating memory units. LSTM has been successfully adopted to several tasks, e.g., speech recognition, language translation and image captioning \cite{cho2015describing,venugopalan2014translating}. Thus, we follow this elegant recipe and choose to extend LSTM to generate the video sentence with semantic content.

Early attempts were proposed \cite{venugopalan2014translating,venugopalan2015sequence,yao2015describing,li2015summarization} to directly connect a visual convolution model to a deep LSTM networks. For example, Venugopalan \textit{et al.} \cite{venugopalan2014translating} translate videos to sentences by directly {concatenating} a deep neural network with a recurrent neural network.
More recently, attention mechanism \cite{GuLLL16} is a standard part of the deep learning toolkit, contributing to impressive results in neural machine translation \cite{luong2015effective}, visual captioning \cite{xu2015show,yao2015describing} and question answering \cite{Yang_2016_CVPR}. Visual attention models for video captioning make use of video frames at every time step, without explicitly considering the semantic attributes of the predicted words. For example, \eat{in Fig.~\ref{fig.framework1}, }some words (i.e., ``man", ``shooting" and ``gun") belong to visual words which have corresponding canonical visual signals, while other words (i.e., ``the", ``a" and ``is") are non-visual words, which require no visual information but language context information~\cite{Lu2016Knowing}. In other words, current visual attention models make use of visual information for generating each work, which is  unnecessary or even misleading. Ideally, video description not only requires modeling and integrating their {sequence dynamic temporal attention information into a natural language but also needs to take into account the relationship between sentence semantics and visual content} \cite{DBLP:journals/corr/GanGHPTGCD16}, which to our knowledge has not been simultaneously considered.

\eat{\begin{figure}[t]
	\centering
	\includegraphics[width=0.8\linewidth]{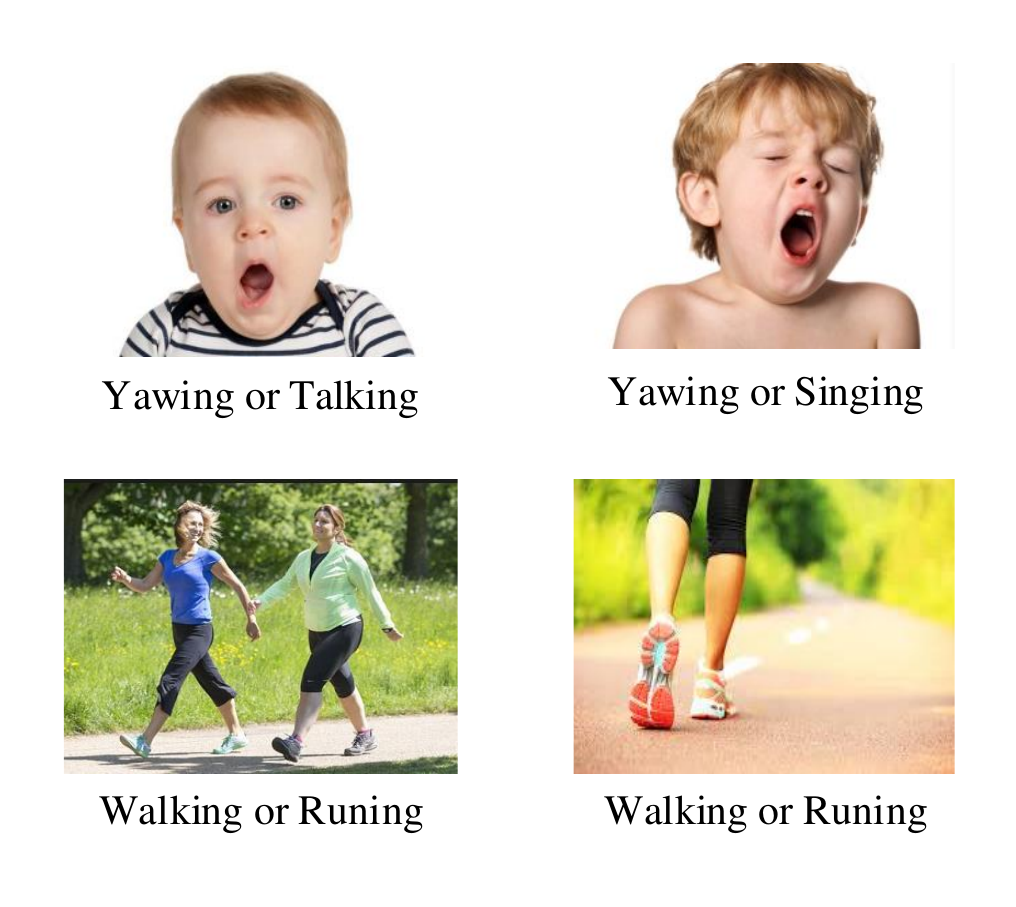}
	\caption{Examples: some actions cannot be identified from a still image or from their appearance alone.}
	\label{fig:DEMO}
\end{figure}}

\eat{Over the past few years, some researchers have proposed to integrate multiple features to improve the performance of various video based applications, such as visual tracking \cite{ICCV7410715}, video action recognition \cite{Simonyan:2014:TCN:2968826.2968890,Duta_2017_CVPR}, 3D object detection \cite{Chen_2017_CVPR} and video retrieval \cite{6553136}. Single visual feature is often insufficient to characterize the video content and each type of feature suffers from some limitations to handle the variance of objects and actions. Intuitively, the multiple features of video clips, each of which reflects the specific information of video data, are complementary to each other. Simultaneously utilizing the multiple features is helpful for precise word generation. For instance, some actions can be identified from a still image from their appearance alone (e.g., shooting). For others, through, individual frames can be ambiguous, and motion cues are necessary \cite{Simonyan:2014:TCN:2968826.2968890,Feichtenhofer_2016_CVPR}, such as, discriminating waking from running, yawning from laughing, or yawning from singing (see Fig.~\ref{fig:DEMO}) . In this paper, we aim to integrate hLSTMat with two-stream hypothesis to further improve video captioning. }

Although the encoder-decoder framework with attention mechanisms has been widely applied to solve other sequence generation tasks, e.g., machine translation \cite{meteor} and image annotation \cite{Genome}. 
In standard sequence generation tasks, an input is encoded to a vector embedding, and then decoded to an output string of words using RNNs, e.g., LSTM.
However, such framework is in essential an one-pass forward process. 
When a model predicts the next word, it can only leverage the generated words but not the future unknown words. To humans, deliberation action is a common behavior in their daily, e.g., reading, writing or understanding an image. During the process, global information of both the past and future are leveraged. Xie \textit{et. al.} \cite{deliberation} designed a deliberation network which included two levels of decoders, and it is proved to be effective for neural machine translation. The first one generates a coarse sentence and corresponding hidden states. The second decoder refines the sentence with deliberation. In the network, the second decoder could leverage the global information of both the past and future parts.
Also, the success of deep neural networks is commonly attributed to the hierarchy that is introduced due to the several layers. Each layer processes some part of the task we wish to solve, and passes it on to the next. However, most of the current visual captioning tasks utilize LSTM with a single layer.

To tackle these issues, \eat{inspired by the attention mechanism for image captioning~\cite{Lu2016Knowing}, }in this paper we propose a unified encoder-decoder framework (see Fig.~\ref{fig.framework0}), named hLSTMat, a Hierarchical LSTMs with adaptive temporal attention model for visual captioning. Specifically, first, in order to extract more meaningful visual features, we adopt a deep neural network to extract region-level or frame-level 2D CNN features for each video or image. Next, we integrate a hierarchical LSTMs consisting of two layers of LSTMs and adaptive attention to decode visual information and language context information to support the generation of sentences for  video and image description. Moreover, the proposed novel adaptive attention mechanism automatically decides whether to rely on visual information or not. % When relying on visual information, the model enforces the gradients from visual information to support visual captioning. Otherwise, the model predicts the words using natural language context without considering visual signals. 

It is worthwhile to highlight the main contributions of this proposed approach:
1) We introduce a novel hLSTMat framework which automatically decides when and where to use visual information, and when and how to adopt the language model to generate the next word for visual captioning.
Specifically, the spatial and temporal  attention is used to decide where to look at visual information and the adaptive attention decides when to rely on language context information.
2) A hierarchical LSTMs is designed to enrich the representation ability of the LSTM. Specifically, when the two LSTMs are connected at each time step, the hierarchical LSTMs can obtain both low-level visual information and high-level language context information. On the other hand, when the two LSTMs are connected sequentially, the second LSTM can be considered as a deliberation process to refine the first LSTM because it is based on the rough global features captured by the first LSTM.
3) Extensive experiments are conducted on mainstream benchmark datasets for both video and image captioning tasks.
Experimental results show that our approach achieves the state-of-the-art performance for most of the evaluation metrics on both tasks. The effect of important components is also well exploited in the ablation study. 

The rest of our paper is organized as follows: Firstly, works about visual captioning are introduced in Sec. \ref{sec.relatedwork}. Then we describe the details of our proposed model hLSTMat in Sec. \ref{sec.methodology}. We first instantiate our hLSTMat for the task of video captioning in Sec. \ref{sec.videoCap}, and show the experimental results for video captioning in Sec. \ref{sec.exp1}. Then we further instantiate our hLSTMat for the task of image captioning in Sec. \ref{sec.imageCap}, and show the experimental results for image captioning in Sec. \ref{sec.exp2}. Finally, we conclude our method in Sec. \ref{sec.conclusion}.

\section{Related Work}
\label{sec.relatedwork}
Image and video captioning, which automatically translates images and videos into natural language sentences to describe their content, is a very important task. Recent years, we have witnessed the significant growth of captioning and previous work is mainly focusing on two aspects: feature extraction and model innovation.

\subsection{Feature Extraction}
Local feature or local image descriptor is at the core of many computer vision tasks, and classical methods such as HOF, HOG and SIFT have been extensively used in various computer vision applications such as image classification, object detection, image retrieval and segmentation. Alone with the booming of handcrafted descriptors in the past decade and the large collected datasets like ImageNet, more and more learning based descriptors appears, such as  AlexNet \cite{krizhevsky2012imagenet}, GoogLenet \cite{szegedy2015going}, VGG \cite{simonyan2014very} and ResNet \cite{he2015deep}. Different from handcrafted descriptors which are mostly driven by intuitions or domain expert knowledge, learning based methods are driven by the large scale annotated datasets and the rapid development of GPU \cite{Tian_2017_CVPR}. To date, deep learning has revolutionized almost the whole computer vision fields and even other areas such as natural language processing and medical image research. Inspired by the success of recurrent neural network (RNN) in neural machine translation and deep convolutional neural network (CNN) in various visual analysis tasks, image captioning is proposed since associating image captioning with machine translation makes sense. They can be placed in the same framework, called Encoder-Decoder framework. As a result, numerous discriminative CNN features have been proposed an adopted in image captioning task to extract appearance features, such as AlexNet \cite{krizhevsky2012imagenet}, GoogLenet \cite{szegedy2015going}, VGG \cite{simonyan2014very} and ResNet \cite{he2015deep}. Compared with learning based image descriptor, video descriptor has not yet seen the substantial gains in performance that have been achieved in other areas by CNNs, e.g., image classification and object detection. C3D \cite{tran2015learning} model is proposed to explore temporal information and good at extracting motion information. More video dataset is required to improve the performance of learning based video descriptor. Therefore, in this paper, we adopt CNNs to extract image/frame appearance feature and C3D model to capture video motion information.

\begin{figure*}[t]
	\centering
	\includegraphics[width=0.65\linewidth]{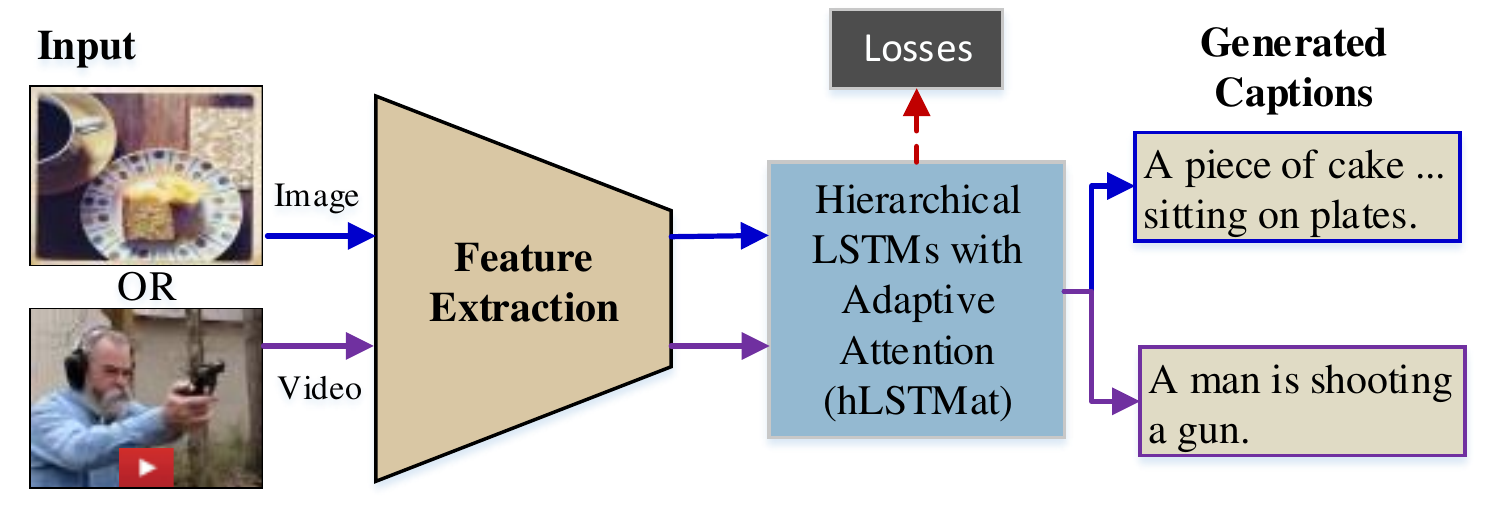}
	\caption{The framework of our proposed hLSTMat for visual captioning. Given an input image or video, an encoder is firstly applied to extract the features. Then hierarchical LSTM with adaptive attention component plays the role of an decoder, by using the hierarchical LSTM to extract different level of information, and an adaptive attention to decide whether to depend on the visual information or the language context information. The losses are defined on the generated captions and the groundtruth to guide the learning of network parameters. Note that when this framework is applied to different visual captioning tasks, i.e., image and video captioning, there are differences in terms of feature extractor, network structure and losses.}
	\label{fig.framework0}
\end{figure*}

\subsection{Model Innovation}
Except improving the image/video captioning with better features, researchers are also focusing on captioning model innovation, which plays an irreplaceable role in improving the performance. 

\textbf{Stage 1: Basic Encoder-Decoder.} At the earlier stage of visual captioning, several models such as \cite{venugopalan2015sequence,vinyals2015show,xu2015jointly} have been proposed by directly bring together previous advances in natural language processing and computer vision. More specifically, semantic representation of an image is captured by a CNN network and then decoded into a caption using various architectures, such as recurrent neural networks.  For example,  Venugopalan \textit{et al.} \cite{venugopalan2015sequence} proposed a S2VT approach, which incorporates a stacked LSTM by firstly reading the visual sequence, comprised of RGB and/or optical flow CNN outputs, and then generating a sequence of words. Oriol \textit{et al.} \cite{vinyals2015show} presented a generative model based on a deep recurrent neural network. This model consists of a vision CNN followed by a language generating RNN, and is trained to maximize the likelihood of the target description sentence given the training image. In \cite{xu2015jointly}, the proposed framework consists of three parts: a compositional semantics language model, a deep video model and a joint embedding model. In the joint embedding model, the distance of the outputs of the deep video model and compositional language model is minimized in the joint space. 

\textbf{Stage 2: Visual Attention.} Later on, researchers found that different regions in images and frames in videos have different weights, and thus various attention mechanisms are introduced to guide captioning models by telling where to look at for sentence generation, such as \cite{pan2015hierarchical,Yu_2016_CVPR,yao2015describing,xu2015show}. Yao \textit{et al.} \cite{yao2015describing} proposed to incorporate both the local dynamic of videos as well as their global temporal structure for describing videos. For simplicity, they were focusing on highlighting only the region having the maximum attention. A Hierarchical
Recurrent Neural Encoder (HRNE) \cite{pan2015hierarchical}, was introduced to generate video representation with emphasis on temporal modeling by applying a LSTM along with attention mechanism to each temporal time step. In \cite{long2016video}, it combines multiple forms of attention for video captioning. Temporal, motion and semantic features are weighted via an attention mechanism.

\textbf{Stage 3: Semantic Attention.} Semantic attention has been proposed in previous work \cite{You_2016_CVPR,yao2016boosting,pan2016video} by adopting attributes or concepts generated by other pre-trained models to enhance captioning performance. Basically, semantic attention is able to attend to a semantically important concepts or attributes or region of interest in an image, and able to weight the relative strength of attention paid on multiple concepts \cite{You_2016_CVPR}. In addition, Yao \textit{et al.} presented Long Short-Term Memory with Attributes (LSTM-A) to integrates attributes into the successful CNNs and RNNs for image captioning. Variant architectures were constructed to feed image features and attributes into RNNs in different ways to explore the mutual but also fuzzy relationship between them. Moreover, Pan \textit{et al.} \cite{pan2016video} proposed Long Short-Term Memory with Transferred Semantic Attributes (LSTM-TSA), which takes advantages of incorporating transferred semantic attributes learnt from images and videos into sequence learning for video captioning.

\textbf{stage 4: Deep Reinforcement Learning based Approaches.} Recently, several researchers have started to utilize reinforcement learning to optimize image captioning \cite{ranzato2015sequence,rennie2016self,liu2016improved,ren2017deep}. To collaboratively generate caption, Ren \textit{et al.} \cite{ren2017deep} incorporated a ``policy network" for generating sentence and a ``value network" for evaluating predicting sentence globally, and they improved the policy and value networks with deep reinforcement learning. Furthermore, Liu  \textit{et al.} \cite{liu2016improved} proposed to improve image captioning via policy gradient optimization of a linear combination of SPICE and CIDEr, while Rennis \textit{et al.} \cite{rennie2016self}  utilized the output of its own test-time inference algorithm to normalize the rewards it experiences.  All the previous methods show that reinforcement learning has the potential to boost image captioning.

\section{Hierarchical LSTM with Adaptive Attention for Visual Captioning}
\label{sec.methodology}
In this section, we first briefly describe how to directly use the basic Long Short-Term Memory (LSTM) as the decoder for visual captioning task. Then we introduce our novel encoder-decoder framework, named Hierarchical LSTM with Adaptive Attention (hLSTMat) (See Fig.~\ref{fig.framework0}) for both video and image captioning.% Finally, we introduce our two-stream adaptive temporal attention mechanism.
% More specifically, a deep CNN encoder is applied to extract features from the input video frames. Then, a hierarchical LSTMs with adaptive temporal attention mechanism is applied to efficiently generate sentences for describing the input video. Detail information is given below.

\subsection{A Basic LSTM for Visual Captioning}
%\textcolor{red}{LSTM is not for encoder, so I rename it}

To date, modeling sequence data with Recurrent Neural Networks (RNNs) has shown great success in the process of machine translation, speech recognition, image and video captioning \cite{chen2014learning,fang2015captions,venugopalan2014translating,venugopalan2015sequence} etc.
Long Short-Term Memory (LSTM) is a variant of RNN to avoid the vanishing gradient problem \cite{bengio1994learning}.
% However, it is still challenging to train a standard RNN due to the vanishing gradient problem \cite{bengio1994learning}. As an update version of the standard RNN, Long Short-Term Memory (LSTM) has sloven this problem by learning patterns with wider range temporal dependencies. 
%To date, using LSTM for both image and video captioning  obtains impressive results. 

\textbf{LSTM Unit.} A basic LSTM unit consists of three gates (input ${\mathbf{i}_t}$, forget ${\mathbf{f}_t}$ and output ${\mathbf{o}_t}$), a single memory cell ${\mathbf{m}_t}$. Specifically, ${\mathbf{i}_t}$ allows incoming signals to alter the state of the memory cell or block it. ${\mathbf{f}_t}$ controls what to be remembered or be forgotten by the cell, and somehow can avoid the gradient from vanishing or exploding when back propagating through time. Finally, ${\mathbf{o}_t}$ allows the state of the memory cell to have an effect on other neurons or prevent it. Basically, the memory cell and gates in a LSTM block are defined as follows:
\begin{equation}
\begin{aligned}
\mathbf{i}_{t} & = \sigma (\mathbf{W}_{i} \mathbf{y}_{t} +  \mathbf{U}_{i} h_{t-1} + \mathbf{b}_{i}) \\
\mathbf{f}_{t} & = \sigma (\mathbf{W}_{f} \mathbf{y}_{t} +  \mathbf{U}_{f} h_{t-1} + \mathbf{b}_{f}) \\
\mathbf{o}_{t} & = \sigma (\mathbf{W}_{o} \mathbf{y}_{t} +  \mathbf{U}_{o} h_{t-1} + \mathbf{b}_{o}) \\
\mathbf{g}_{t} & = \phi (\mathbf{W}_{g} \mathbf{y}_{t} +  \mathbf{U}_{g} h_{t-1} + \mathbf{b}_{g}) \\
\mathbf{m}_{t} & = \mathbf{f}_{t} \odot \mathbf{m}_{t-1}+ 
\mathbf{i}_{t} \odot \mathbf{g}_{t}         \\
\mathbf{h}_{t} & = \mathbf{o}_{t}\odot \phi(\mathbf{m}_{t})
\end{aligned}
\label{eq.lstm}
\end{equation}
where the weight matrices $\mathbf{W}$, $\mathbf{U}$, and $\mathbf{b}$ are parameters to be learned. $\mathbf{y}_{t}$ represents the input vector for the LSTM unit at each time $t$. $\sigma$ represents the logistic sigmoid non-linear activation function mapping real numbers to $(0,1)$, and it can be thought as knobs that LSTM learns to selectively forget its memory or accept the current input. $\phi$ denotes the hyperbolic tangent function tanh. $\odot$ is the element-wise product with the gate value. For convenience, we denote $\mathbf{h}_{t},\mathbf{m}_{t}=\mathrm{LSTM}(\mathbf{y}_{t}, \mathbf{h}_{t-1}, \mathbf{m}_{t-1})$ as the computation function for updating the LSTM internal state.

\textbf{Visual Captioning.} Given an input video or image $\mathbf{x}$, an encoder network $\phi_{E}$ encodes it into a representation space:
\begin{equation}
%\mathbf{V}=\{\mathbf{v}_{1}, \cdots, \mathbf{v}_{n}\} = \phi_{E}(\mathbf{x}).
\mathbf{V} = \phi_{E}(\mathbf{x}).
\end{equation}
where $\phi_{E}$ usually denotes a CNN neural network. Here, LSTM is chosen as a decoder network $\phi_{D}$ to model $\mathbf{V}$ to generate a description $\mathbf{z}=\{z_{1}, \cdots, z_{T}\}$ for $\mathbf{x}$, where $T$ is the description length. In addition, the LSTM unit updates its internal state $\mathbf{h}_{t}$ and the $t$-th word $z_{t}$ based on its previous internal state $\mathbf{h}_{t-1}$, the previous output $y_{t}$ and the representation $\mathbf{V}$:
\begin{equation*}
\left( \mathbf{h}_{t},z_{t} \right)
= \phi_{D}(\mathbf{h}_{t-1},y_{t},\mathbf{V})
\end{equation*}
In test stage, the previous output $\mathbf{z}_{t-1}$ is the current input vector $\mathbf{y}_{t}$. And the initial $\mathbf{y}_{0}$ represents the begin of the sentence $ (BOS)$.

In addition, the LSTM updates its internal state recursively until the end-of-sentence tag is generated. For simplicity, we named this simple method as basic-LSTM.

\begin{figure*}[t]
	\centering
	\includegraphics[width=0.85\linewidth]{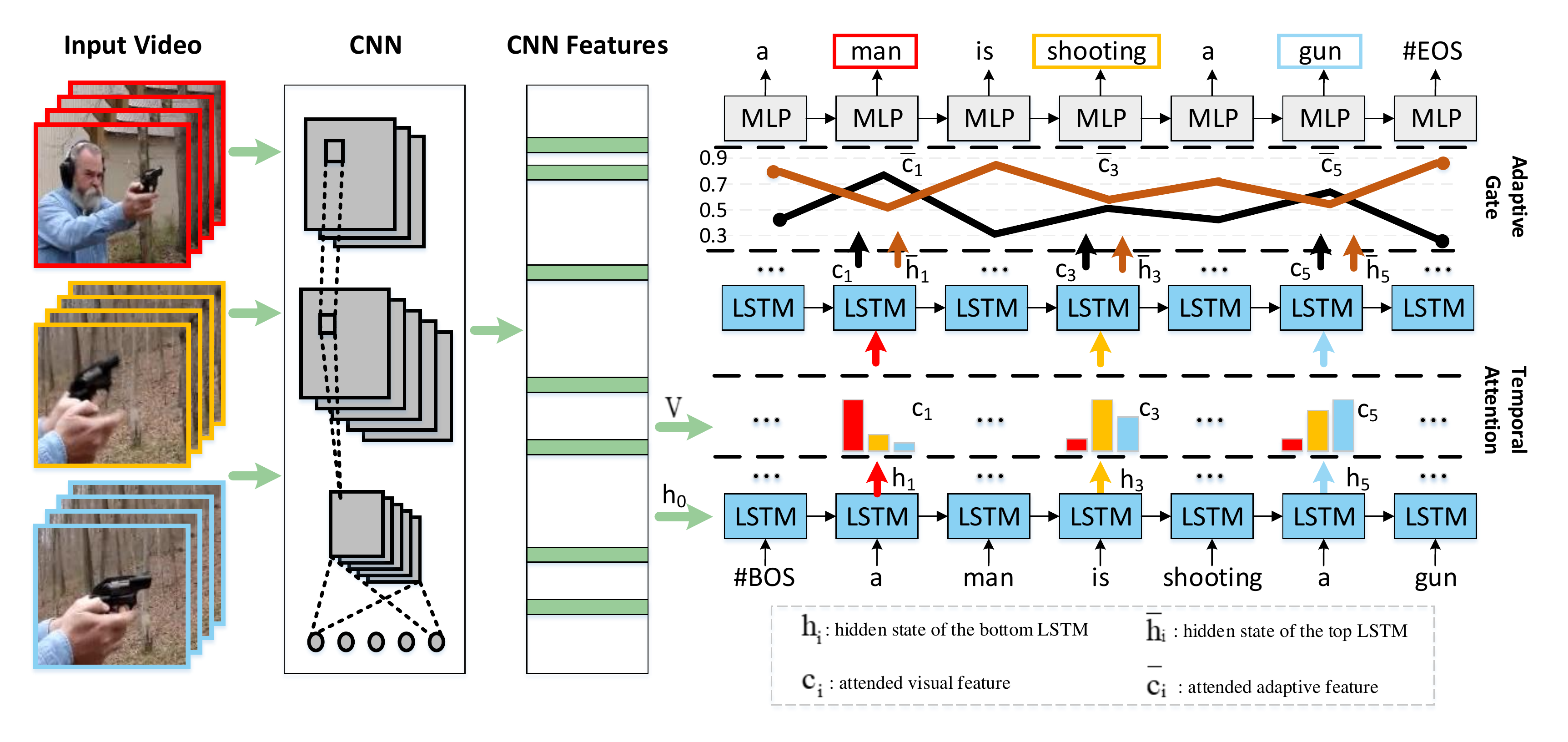}
	\caption{{The framework of our proposed method hLSTMat. To illustrate the effectiveness of our hLSTMat, each generated visual words (i.e., ``man", ``shooting" or ``gun") is generated with visual information extracting from a set of specific frames. For instance, ``man" is marked with ``red", this indicates it is generated by using the frames marked with red bounding boxes, ``shooting" is generated replying on the frames marked with ``orange". Other non-visual words such as ``a" and ``is" are relying on the language model.}}
	\label{fig.framework1}
\end{figure*}

\subsection{Hierarchical LSTMs with Adaptive Temporal Attention for Visual Captioning}
\label{method0}
In this subsection, we introduce our hLSTMat framework, which consists of three major components: 1) an encoder; 2) an attention based hierarchical LSTM decoder; and 3) the losses.

\subsubsection{CNN Encoders} The goal of an encoder is to compute features that are compact and representative and can capture the most related visual information for the decoder. Thanks to the rapid development of CNNs, a great success has been made in large scale image recognition task \cite{he2015deep}, object detection \cite{ren2015faster} and visual captioning \cite{venugopalan2014translating}. High-level features can been extracted from upper or intermediate layers of a deep CNN network. Therefore, a set of well-tested CNN networks, such as the ResNet-152 model \cite{he2015deep} which has achieved the best performance in ImageNet Large Scale Visual Recognition Challenge, can be used as a candidate encoder for our framework. 
Depending on the type of input, different types of features can be extracted. For example, we can extract frame-level features for videos and we can extract region-level features for images.
The details of feature extraction are task-specific, and will be given in each specific task in Section \ref{sec.videoCap} and Section \ref{sec.imageCap}.

\subsubsection{hLSTMat Decoder}
In the basic model, a LSTM is used as the decoder to predict the captions. The vanilla LSTM model is comprised of a single hidden LSTM layer followed by a standard feedforward output layer.
However, the success of deep neural networks is commonly attributed to the hierarchy that is introduced due to the several layers \cite{NIPS2013_5166}. Therefore, we proposed a hierarchical LSTM based decoder. It can be seen as a processing pipeline, in which each layer solves a part of the task before passing it on to the next, until finally the last layer provides the output.

Also, we define an adaptive attention layer based on the hierarchical LSTM for deciding whether to depend on the visual information or the language context information.

\subsubsection{Losses}
After the top MLP layer predicts the probability distribution of each word in the vocabulary, maximum likelihood estimation (MLE) loss is usually utilized to guide the learning of parameters:
\begin{equation}
\ell_{MLE} = - \sum_{t=1}^{T} log P(z_{t}|z_{<t}, \mathbf{V},\Theta)
\label{eq:mle}
\end{equation}
{where $z_{t}$ is the $t$-th word in the ground truth caption, $z_{<t}$ are the first $t$-$1$ words in the ground truth caption}, $T$ denotes the total number of words in sentence, $\mathbf{V}$ denotes the features of the corresponding input, and $\Theta$ are model parameters. Therefore, Eq.~\ref{eq:mle} is regarded as the MLE loss function to optimize the model.

There are also other types of losses for visual captioning. We consider our method introduced above as ``agent'' to interact with external environment (i.e., words, visual features),  and $\Theta$ as the policy to conduct an action to predict a word. After the whole caption is generated, the agent observers a reward. Since CIDEr is proposed to evaluate the quality of visual captioning model. We can design our reward functions using CIDEr, and the loss based on deep reinforcement learning can be defined.

\section{hLSTMat for Video Captioning}
\label{sec.videoCap}
In this subsection, we instantiate our hLSTMat and apply it to the task of video captioning. As shown in Fig.~\ref{fig.framework1}, there are three major components: 1) a CNN Encoder, 2) an attention based hierarchical LSTM decoder and 3) the losses. We give the details for each component below. 

\subsection{CNN Encoders} 
%The goal of an encoder is to compute feature vectors that are compact and representative and can capture the most related visual information for the decoder. Thanks to the rapid development of deep convolutional neural networks (CNNs), which have made a great success in large scale image recognition task \cite{he2015deep}, object detection \cite{ren2015faster} and visual captioning \cite{venugopalan2014translating}. High-level features can been extracted from upper or intermediate layers of a deep CNN network. Therefore, a set of well-tested CNN networks, such as the ResNet-152 model \cite{he2015deep} which has achieved the best performance in ImageNet Large Scale Visual Recognition Challenge, can be used as a candidate encoder for our framework. %With a CNN architecture, we can apply it to each frame to extract representative frame-level features.

Inspired by \cite{yao2015describing}, we preprocess each video clip by selecting equally-spaced 28 frames out of the first 360 frames and then feeding them into a CNN network proposed in \cite{he2015deep}. Thus, for each selected frame we obtain a 2,048-D feature vector, which are extracted from the $pool5$ layer. 
 For simplicity, given a video $\textbf{x}$, we extract $L$=$28$ frame-level features $\textbf{V}$, represented as $\textbf{V} = \left\{ {{\textbf{v}_1}, \cdots ,{\textbf{v}_L}} \right\}$.
 
{Motion features catches the long range dependencies of video, which is quite essential for video information. The C3D motion net reads 16 adjacent frames as a segment and catches a 4096-D motion feature vector as output. For each video, we attain 10 segment features to represent the whole video.}

\begin{figure}[t]
	\centering
	\includegraphics[width=0.99\linewidth]{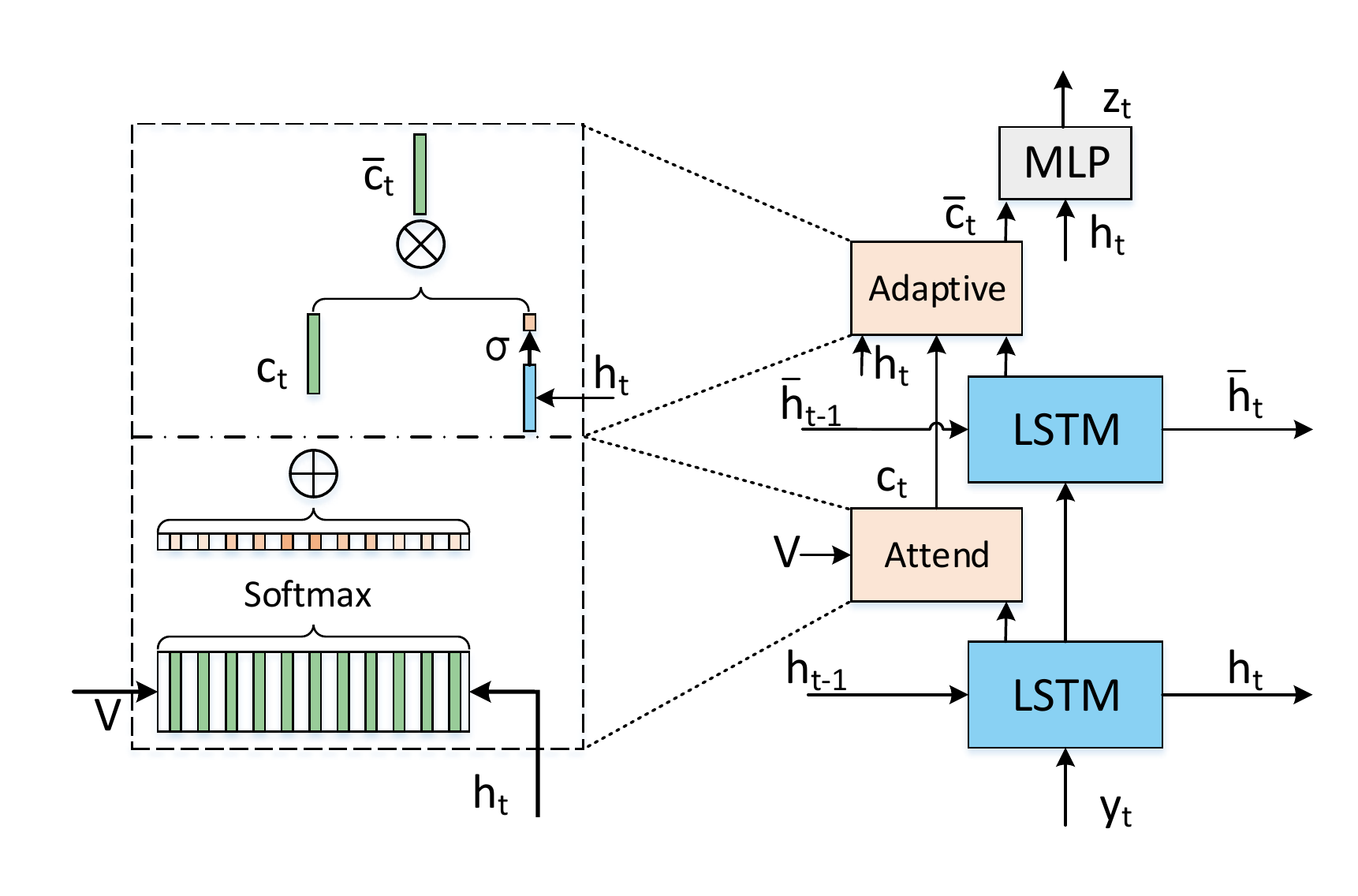}
	\caption{{An illustration of the proposed method generating the t-th target word $z_t$ given a video.}}
	\label{fig:our_attention}
\end{figure}

\subsection{Attention based Hierarchical Decoder}

Our decoder (see Fig.~\ref{fig:our_attention}) integrates two LSTMs. 
The bottom LSTM layer is used to efficiently decode visual features, and the top LSTM is focusing on mining deep language context information for video captioning.
We also incorporate two attention mechanisms into our framework. A temporal attention is to guide which frame to look, while the adaptive temporal attention is proposed to decide when to use visual information and when to use sentence context information.
The top MLP layer is to predict the probability distribution of each word in the vocabulary.

%LSTM has been proven successfully in modeling sequence data and recently attention based LSTM is introduced for video captioning. 
Unlike vanilla LSTM decoder, which performs mean pooling over 2D features across each video to form a fixed-dimension representation, attention based LSTM decoder is focusing on a subset of consecutive frames to form a fixed-dimensional representation at each time $t$. 
\begin{itemize}
	\item{ Bottom LSTM Layer. For the bottom LSTM layer, the updated internal hidden state depends on the current word $y_{t}$, previous hidden state $\mathbf{h}_{t-1}$ and memory state $\mathbf{m}_{t-1}$:
		\begin{equation}
		\begin{aligned}
		\mathbf{h}_{0},\mathbf{m}_{0} & =  \left[\mathbf{W^{ih}}; \mathbf{W^{ic}} \right]  Mean(\{\mathbf{v_{i}}\})  \\
		\mathbf{h}_{t},\mathbf{m}_{t} &= LSTM(\mathbf{y}_{t}, \mathbf{h}_{t-1}, \mathbf{m}_{t-1}) \\
		\end{aligned}
		\end{equation}
		where $\mathbf{y}_{t} = \mathbf{E}[y_{t}]$ denotes a word feature of a single word $y_{t}$. $Mean(\cdot)$ denotes a mean pooling of the given feature set $\mathbf{v_{i}}$. $\mathbf{W^{ih}}$ and $\mathbf{W^{ic}}$ are parameters that need to be learned.
	}
	
	\item{Top LSTM Layer. For the top LSTM, it takes the output of the bottom LSTM unit output $\mathbf{h}_{t}$, previous hidden state $\mathbf{\bar{h}}_{t-1}$ and the memory state $\mathbf{\bar{m}}_{t-1}$ as inputs to obtain the hidden state $\mathbf{\bar{h}}_{t}$ at time $t$, and it can be defined as below: 
		\begin{equation}
		\begin{aligned}
		\mathbf{\bar{h}}_{0}&=0, \mathbf{\bar{m}}_{0}=0 \\
		\mathbf{\bar{h}}_{t},\mathbf{\bar{m}}_{t}  & = LSTM( \mathbf{h}_{t}, \mathbf{\bar{h}}_{t-1}, \mathbf{\bar{m}}_{t-1}) \\
		\end{aligned}
		\end{equation}
	}
	\item{Attention Layers.
		In addition, for attention based LSTM, context vector is in general an important factor, since it provides  meaningful visual evidence for caption generation \cite{yao2015describing}. In order to efficiently adjust the choose of visual information or sentence context information for caption generation, we defined an adaptive temporal context vector $\mathbf{\bar{c}}_{t} $ and a temporal context vector $\mathbf{c}_{t}$ at time $t$. See below: 
		\begin{align}
		\begin{aligned}
		\mathbf{\bar{c}}_{t}   & = \psi(\mathbf{h}_{t}, \mathbf{\bar{h}}_{t}, \mathbf{c}_{t}),~~
		\mathbf{c}_{t} & = \varphi(\mathbf{h}_{t}, \mathbf{V}) \\
		\end{aligned}
		\label{Eq:cv}
		\end{align}
		where $\psi$ denotes the function of our adaptive gate, while $\varphi$ denotes the function of our temporal attention model. Moreover, $\mathbf{\bar{c}}_{t}$ denotes the final context vector through our adaptive gate, and $\mathbf{c}_{t}$ represents intermediate vectors calculated by our temporal attention model.
		These two attention layers will be described in details in Sec.~\ref{sec.sub.tem.att} and Sec.~\ref{sec.sub.adj.tem.att}.
	}
	
	\item MLP layer. To output a symbol $z_{t}$, a probability distribution over a set of possible words is obtained using $\mathbf{h}_{t}$ and our adaptive temporal attention vector $\mathbf{\bar{c}}_{t} $:
	\begin{equation}
	\mathbf{p}_{t} = softmax \left( \mathbf{U}_{p} \phi(\mathbf{W}_{p}[\mathbf{h}_{t};\mathbf{\bar{c}}_{t}]+\mathbf{b}_{p}) + \mathbf{d}  \right)
	\end{equation}
	where $\mathbf{U}_{p}$, $\mathbf{W}_{p}$, $\mathbf{b}_{p}$ and $\mathbf{d}$ are parameters to be learned. Next, we can interpret the output of the softmax layer $\mathbf{p}_{t}$ as a probability distribution over words:
	\begin{equation}
	P(z_{t}|z_{<t}, \mathbf{V},\Theta)
	\end{equation}
	where $\mathbf{V}$ denotes the features of the corresponding input video, and $\Theta$ are model parameters.
\end{itemize}

\subsection{Losses}

To learn $\Theta$ in our modal, we minimize the MLE loss defined in Eq.~\ref{eq:mle}.
After the parameters are learned, we choose BeamSearch \cite{vinyals2015show} method to generate sentences for videos, which iteratively considers the set of the $k$ best sentences up to time $t$ as candidates to generate sentence of time $t+1$, and keeps only best k results of them. Finally, we approximates $D=argmax_{D^{'}}Pr(D^{'}|X)$ as our best generated description.
In our entire experiment, we set the beam size of BeamSearch as 5.

\subsection{Temporal Attention Model}

\label{sec.sub.tem.att}
\eat{{Our temporal attention and the difference between normal and ours}}
\eat{
	\begin{figure}[t]
		\centering
		\includegraphics[width=0.85\linewidth]{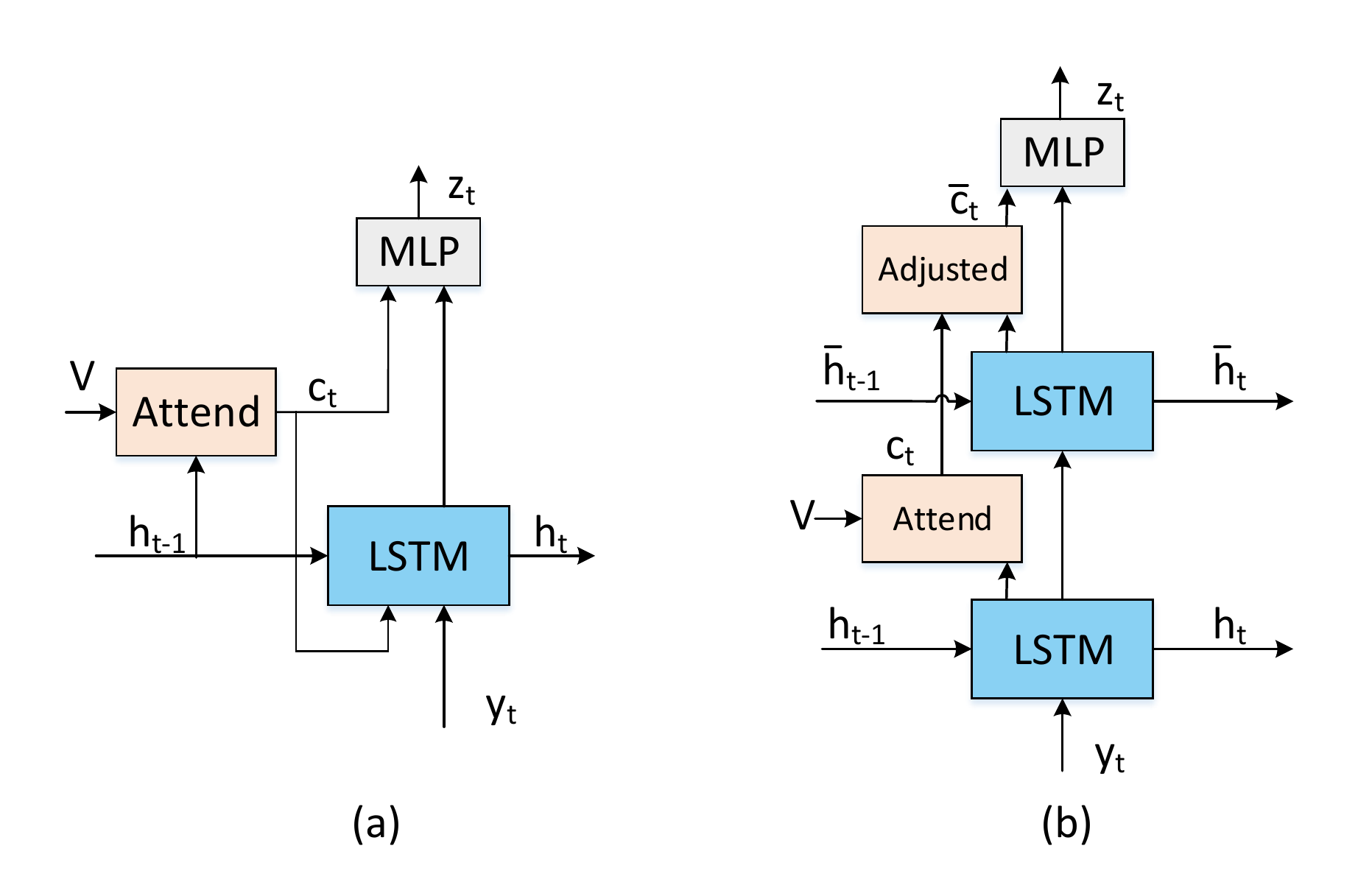}
		\caption{An illustration of soft attention model (a) and our proposed temporal attention model (b).}
		\label{fig:soft_attention}
	\end{figure}
}

As mentioned above, temporal context vector $\mathbf{c}_{t}$ is an important factor in encoder-decoder framework. To deal with the variability of the length of videos, a simple strategy \cite{venugopalan2014translating} is used to compute the average of features across a video, and this generated feature is used as input to the model at each time step:
\begin{align}
\mathbf{c}_{t}  = \frac{1}{L} \sum_{l=1}^{L} \mathbf{v}_{l}
\label{eq:meanpool1}
\end{align}

However, this strategy effectively collapses all frame-level features into a single vector, neglecting the inherent temporal structure and leading to the loss of information. Instead of using a simple average strategy (see Eq.~\ref{eq:meanpool1}), we wish to take the dynamic weight sum of the temporal feature vectors according to attention weights $\alpha_{t}^{i}$, which are calculated by a soft attention. For each $\mathbf{v}_{i}$ at $t$ time step, we use the follow function to calculate $\mathbf{c}_{t}$:
\begin{equation}
\mathbf{c}_{t}  = \frac{1}{L} \sum_{l=1}^{L} \alpha_{t}^{l} \mathbf{v}_{l}
\end{equation}
where at $t$ time step $\sum_{l=1}^{L} \alpha_{t}^{l} =1$.

In this paper, we integrate two LSTM layers, a novel temporal attention model for computing the context vector  $\mathbf{c}_{t}$ in Eq. \ref{Eq:cv} proposed in our framework. Given a set of video features $ \mathbf{V} $ and the current hidden state of the bottom layer LSTM $\mathbf{h}_{t}$, we feed them into a single neural network layer, and it returns an unnormalized relevant scores $\varepsilon_{t}$. Finally, a softmax function is applied to generate the attention distribute over the $L$ frames of the video:
\begin{equation}
\begin{aligned}
\varepsilon_{t} & =  \mathbf{\text{W}}^{\rm T} \left( \mathbf{W}_{a} \mathbf{h}_{t} + \mathbf{U}_{a} \mathbf{V} + \mathbf{b}_{a} \right) \\
\alpha_{t} & = softmax(\varepsilon_{t}) \\
\end{aligned}
\label{Eq.softatt}
\end{equation}
where $\mathbf{w}^{T}$, $\mathbf{W}_{a}$, $\mathbf{U}_{a}$ and $\mathbf{b}_{a}$ are parameters to be learned. $\alpha_{t} \in \mathbb{R}^{n}$ is the attention weight which quantifies the relevance of features in $\mathbf{V}$.
Different from \cite{yao2015describing}, we utilize the current hidden state instead of previous hidden state $\mathbf{h}_{t}$ generated by the first LSTM layer to obtain the context vector $\mathbf{c}_{t}$, which focuses on salient feature in the video.

\subsection{Spatial Attention Model}
{As temporal attention mentions above, different frames play different roles when generating caption. It is intuitive to think that different regions also have different weights in video captioning. Frame-level image feature extract the global feature of the whole frame. Compared with global feature, spatial features of the image contains more details in the image and we use attention mechanism to select important regions in captioning. {For each frame of video, we extract $res5c$ layer from pre-trained ResNet model which denotes the region feature of video frame.} For each step, we can calculate the spatial context vector using current hidden state $\mathbf{h_t}$ as:}
\begin{equation}
\mathbf{c}_{t}  = \frac{1}{N} \sum_{n=1}^{N} \alpha_{t}^{n} \mathbf{r}_{n}
\end{equation}
where $N$ is the number of regions, $\mathbf{r}_i$ is the spatial feature of video. Similar to temporal attention, a simple softmax function is used to generate the spatial attention map as Eq.~\ref{Eq.softatt}. However, spatial attention model use region feature $R$ instead of frame level feature.

\begin{equation}
\begin{aligned}
\varepsilon_{t} & =  \mathbf{\text{W}}^{\rm T} tanh \left( \mathbf{W}_{a} \mathbf{h}_{t} + \mathbf{U}_{a} \mathbf{R} + \mathbf{b}_{a} \right) \\
\alpha_{t} & = softmax(\varepsilon_{t}) \\
\end{aligned}
\end{equation}
R denotes spatial feature extracted from convolution neural network, which represents region feature of the image.

\begin{figure*}[t]
	\centering
	\includegraphics[width=1\linewidth]{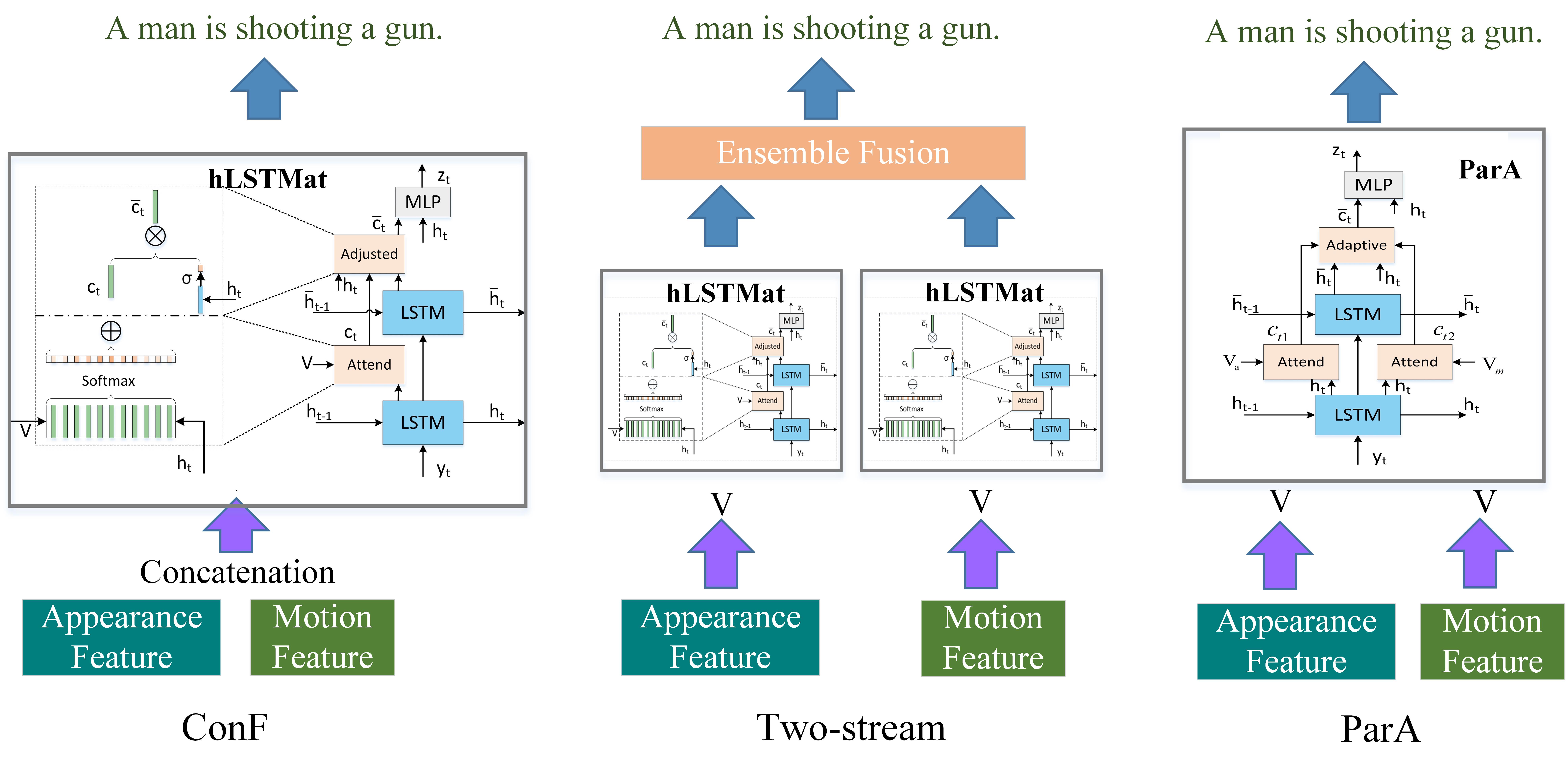}
	\caption{We construct variants of architectures of spatial-temporal hLSTMat. The left indicates the hLSTMat with concatenation fusion (ConF). The middle one represents the Two-stream hLSTMat Networks (Two-stream). The right one illustrates the architectures of parallel adaptive temporal attention model (ParA). }
	\label{fig:fusion}
\end{figure*}

\subsection{Adaptive Temporal Attention Model}
\label{sec.sub.adj.tem.att}

%To date, most decoders of existing methods force the attention mechanism to be active for every generated words including both non-visual words (e.g. ``the" and ``of") that may seems visual but often can be predicted just from the language model, and visual words (e.g., ``man" and ``walking"). Thus, in spite of temporal attention model has achieved better performance for video captioning, they can not determine when to rely on visual information and when to rely on language context information.

\eat{\begin{figure}[h]
	\centering
	\includegraphics[width=0.8\linewidth]{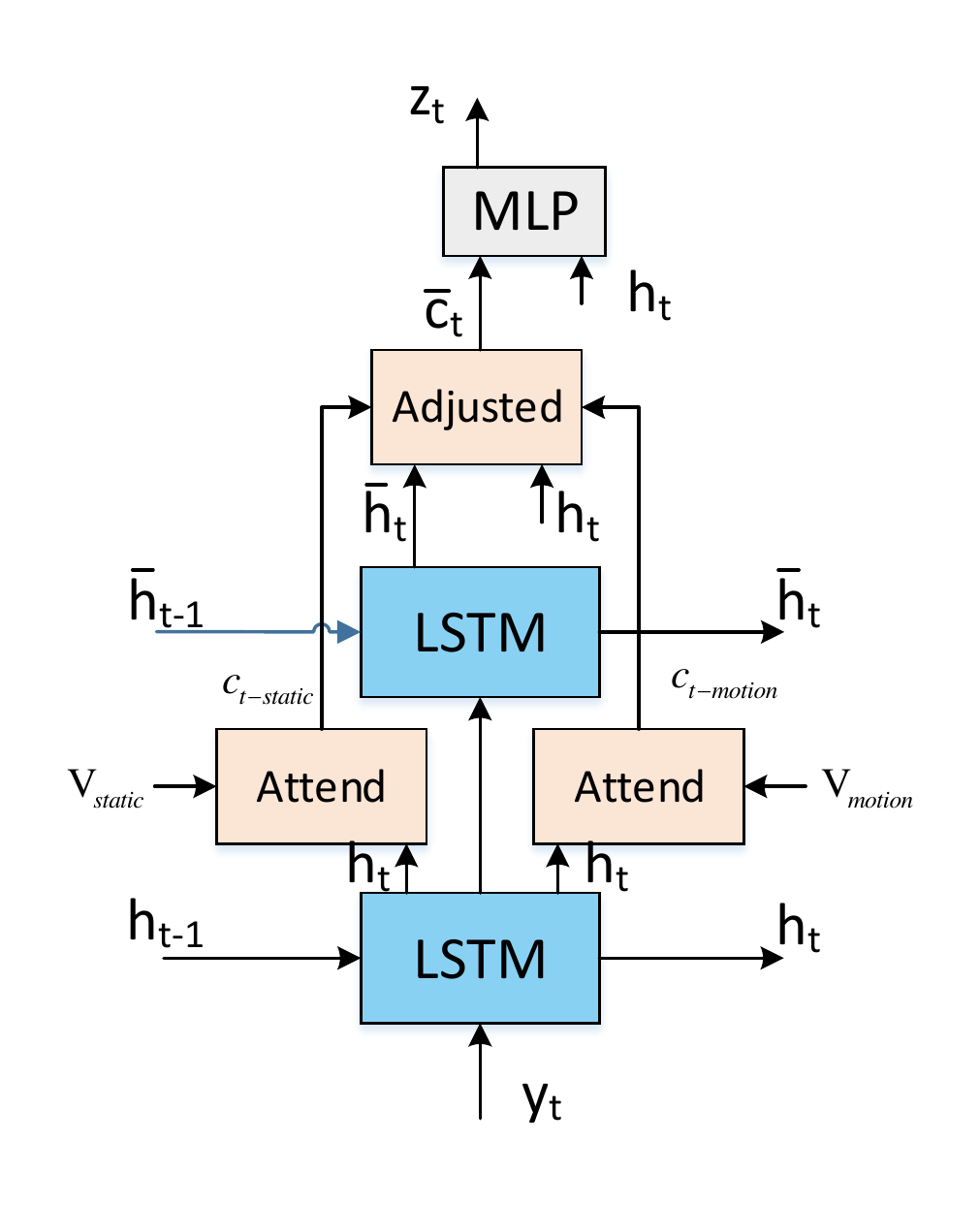}
	\caption{An illustration of our parallel adaptive temporal attention model. }
	\label{fig:multi_feature}
\end{figure}}

In this paper we propose an adaptive temporal attention model to compute a context vector $\mathbf{\bar{c}}_{t}$ in Eq.~\ref{Eq:cv}, shown in Fig. \ref{fig:our_attention}, to encourage that a decoder uses nearly no visual information from video frames to predict the non-visual words, and use the most related visual information to predict visual words. In our hierarchical LSTM network, the hidden state in the bottom LSTM layer is a latent representation of what the decoder already knows. With the hidden state $\mathbf{h}_{t}$, we extend our temporal attention model, and propose an adaptive model that is able to determine whether it needs to attend the video to predict the next word. In addition, a sigmoid function is applied to the hidden state $\mathbf{h}_{t}$ to further filter visual information.
\begin{equation}
\begin{aligned}
\mathbf{\bar{c}}_{t} & = \beta_{t} \mathbf{c}_{t} + (1-\beta_{t})\mathbf{\bar{h}}_{t} \\
\beta_{t} & = sigmoid(\mathbf{W}_s \mathbf{h}_{t})
\end{aligned}
\label{eq.ct}
\end{equation}
where $\mathbf{W}_s$ denotes the parameters to be learned and $\beta_{t}$ is adaptive gate at each time $t$. In our adaptive temporal attention model, $\beta_{t}$ is projected into the range of $[0, 1]$. When $\beta_{t}=1$, it indicates that full visual information is considered, while when $\beta_{t}=0$ it indicates that no visual information is considered to generate the next word.

\subsection{Multiple Features based hLSTMat}
\label{sec.mulLSTM}
%Our proposed hLSTMat encourages the decoder to use language context information to predict the non-visual words and choose the most most related visual information to predict visual words. 
Some visual words (e.g., ``gun'' and ``dog'') and verbs (e.g., ``shooting'' and ``eating'') can be identified from a still image from the appearance alone, while others, (e.g., distinguishing ``yawing'' from ``singing'' or ``walking'' from ``running'') require motion cues as well. The hLSTMat using single appearance feature cannot accurately generate video captions. In this section, we consider three architectures for fusing appearance and motion features.

\textbf{hLSTMat with concatenation fusion (ConF).} In order to make use of two types of features (i.e., appearance and motion cues), a concatenation fusion is introduced by firstly concatenating appearance and motion features to form a new video feature, see Fig.~\ref{fig:fusion}. In other words, the hLSTMat takes an integrated video feature as input to generate video descriptions. 

\textbf{Two-stream hLSTMat Networks (Two-stream).} Two-stream architecture has been widely used in video recognition, and it can achieve very good performance \cite{Simonyan:2014:TCN:2968826.2968890}. Motivated by this, we utilize two different pre-trained hLSTMat models. Each hLSTMat model is trained by taking one type of video feature as the input. As a result, each model separately generates a distribution of the words. For two models, their generated distributions are represented as $\mathbf{p}_{t1}$ and $\mathbf{p}_{t2}$, respectively. To predict the word at time $t$, it uses the following function Eq.~\ref{dis_fusion} to calculate the final distribution $\mathbf{p}_{t}$:
\begin{equation}
\mathbf{p}_{t} = \frac{1}{2}(\mathbf{p}_{t1} + \mathbf{p}_{t2})
\label{dis_fusion}
\end{equation}
Here, we devise our video captioning architecture accordingly, dividing it into two streams, as shown in Fig.~\ref{fig:fusion}. Each stream is implemented using a hLSTMat network.

\begin{table}[]
	\centering
	\caption{The operation modules and their number used in our three proposed models.}
	\begin{tabular}{c||c|c|c|c}
		\hline
		model      & Bot LSTM    & Attention    & Top LSTM & Adap ATT     \\ \hline 
		hLSTMat    & 1           & 1            & 1        & 1            \\ \hline
		ParA       & 1           & 2            & 1        & 1            \\ \hline
		Two Stream & 2           & 2            & 2        & 1            \\ \hline
	\end{tabular}
	\label{tab.modules}
\end{table}

\textbf{Parallel Adaptive Temporal Attention Model (ParA).}
Two-stream hLSTMat Networks fuse the distribution scores using averaging, and the experimental results show that two-stream hLSTMat significantly outperforms hLSTMat. {However, a limitation of two-stream hLSTMat is that it doubles the training time and the number of learning parameters.} In order to solve this problem, we propose a parallel adaptive temporal attention model, which enables hLSTMat to make use of complementary information for word prediction. {The opreation modules, which we use in our hLSTMat, ParA and Two-stream model, are listed as Tab. \ref{tab.modules}. It is obvious that Two-stream model uses double training parameters compared with hLSTMat and there are two attention modules in ParA to deal with two kinds of features. From this table, both of Two-stream and ParA use two attention modules. But for other modules, Two-stream model uses double parameter compared with ConF.} The model is shown in Fig.~\ref{fig:fusion}.

As illustrated in Fig.~\ref{fig:fusion}, appearance ($\mathbf{V}_{static}$) and motion ($\mathbf{V}_{motion}$) features are sent to two parallel \textit{attend} components to extract meaningful appearance and motion evidences. Next, we propose an adaptive temporal model to compute the context vector, which is defined as $\mathbf{\bar{c}}_{t}$. It is a mixture of the video spatial and temporal features as well as the language context information. Mathematically, the ParA model is defined as follows:
\begin{equation}
\label{eq:multi_weights}
\begin{aligned}
\mathbf{\bar{c}}_{t}  = \beta_{t1}& \mathbf{c}_{t1}  + \beta_{t2} \mathbf{c}_{t2} + \beta_{t3}\mathbf{\bar{h}}_{t} \\
[\beta_{t1}, \beta_{t2}, \beta_{t3}] & = softmax(\mathbf{W}_s \mathbf{h}_{t})
\end{aligned}
\end{equation}
where $\beta_{t1}$, $\beta_{t2}$, $\beta_{t3}$ are the new sentinel gates at time $t$. $\mathbf{c}_{t1}$ and $\mathbf{c}_{t2}$ represents the ``appearance sentinel'' vector and the ``motion sentinel'' vector, respectively. If $\beta_{t1}+\beta_{t2}=1$, then $\beta_{t3}=0$. This implies that no visual information is considered, and only the evidence of appearance and motion are considered. In other words, the model can predict visual words and non-visual words on the basis of weight of appearance, motion and language context information. In addition, all the sentinel gates are computed by a softmax, where $\mathbf{W}_s$ are parameters to be learned.  

In terms of initialization of $\mathbf{h}_{0}$, we define it as follows:  
\begin{equation}
\label{multi_init}
\begin{aligned}
\mathbf{h}_{0},\mathbf{m}_{0} & =  \left[\mathbf{W^{ih}}; \mathbf{W^{ic}} \right]  Mean(\{\mathbf{v_{static},v_{motion}}\})  
\end{aligned}
\end{equation}
where $\mathbf{W^{ih}}$ and $\mathbf{W^{ic}}$ are parameters that need to be learned. Note that this model can be easily extended to support multiple features.

\section{Experiments for Video Captioning}
\label{sec.exp1}
We evaluate our algorithm on the task of video captioning. Specifically, we firstly study the influence of CNN encoders. Secondly, we explore the effectiveness of the proposed components. Next, we compare our results with the state-of-the-art methods on three benchmark datasets.

\subsection{Datasets}
We consider three publicly available datasets that have been widely used in previous work.

\textbf{The Microsoft Video Description Corpus (MSVD).}  This video corpus consists of 1,970 short video clips, approximately 80,000 description pairs and about 16,000 vocabulary words~\cite{chen2011collecting}. Following \cite{yao2015describing,venugopalan2015sequence}, we split the dataset into training, validation and testing set with 1,200, 100 and 670 videos, respectively.

\textbf{MSR Video to Text (MSR-VTT).} In 2016, Xu \textit{et al.} \cite{xu2016msr} proposed the largest video benchmark for video understanding, and especially for video captioning. This dataset contains 10,000 web video clips, and each clip is annotated with approximately 20 natural language sentences. In addition, it covers the most comprehensive categorizes (i.e., 20 categories) and a wide variety of visual content, and contains 200,000 clip-sentence pairs. 
%To date, it represents the largest dataset in terms of vocabulary and sentence for video caption task. Following in \cite{xu2016msr}, this dataset is split according to 65\%:30\%:5\%, corresponding to 6,513, 2,990 and 497 clips in the training, testing and validation sets, respectively. 

\textbf{Large Scale Movie Description Challenge (LSMDC).} It is a dataset of movies with aligned descriptions sourced from movie scripts. Also, it is based on the two previous datasets: the MPII Movie Description dataset (MPII-MD \cite{rohrbach2015dataset}) and the Montreal Video Annotation Dataset (M-VAD \cite{torabi2015using}). In total, this dataset contains a parallel corpus of 118,114 sentences and 118,081 video clips from 202 movies. In our experiments, it is divided into training, validation, public test, and blind test set, and each contains 10,1079, 7,408, 10,053, and 9,578 videos, respectively.

\subsection{Implementation Details }
\subsubsection{Preprocessing} For MSVD dataset, we convert all descriptions to lower cases, and then use wordpunct\_tokenizer method from the NLTK toolbox to tokenize sentences and remove punctuations. Therefore, it yields a vocabulary of 13,010 in size for the training split. For MSR-VTT dataset, captions have been tokenized, thus we directly split descriptions using blank space, thus it yields a vocabulary of 23,662 in size for training split. 
For LSMDC, vocabulary is composed of 25,610 words, which are extracted from captions by the NLTK toolbox.
{Inspired by \cite{yao2015describing}, we preprocess each video clip by selecting equally-spaced 28 frames out of the first 360 frames and then feeding them into a CNN network proposed in \cite{he2015deep}. Thus, for each selected frame we obtain a 2,048-D feature vector, which are extracted from the $pool5$ layer. As for spatial feature, we extract region features from $res5c$ layer.}

\subsubsection{Training details} In the training phase, in order to deal with sentences with arbitrary length, we add a begin-of-sentence tag $<$BOS$>$ to start each sentence and an end-of-sentence tag $<$EOS$>$ to end each sentence. In the testing phase, we input $<$BOS$>$ tag into our attention-based hierarchical LSTM to trigger video description generation process. For each word generation, we choose the word with the maximum probability and stop until we reach $<$EOS$>$.

In addition, all the LSTM unit sizes are set as 512 and the word embedding size is set as 512, empirically. Our objective function Eq.~\ref{eq:mle} is optimized over the whole training video-sentence pairs with mini-batch 64 in size of MSVD 256 in size of MSR-VTT and LSMDC. We adopt adadelta \cite{zeiler2012adadelta}, which is an adaptive learning rate approach, to optimize our loss function. In addition, we utilize dropout regularization with the rate of 0.5 in all layers and clip gradients element wise at 10. We stop training our model until 500 epochs are reached, or until the evaluation metric does not improve on the validation set at the patience of 20.

\subsubsection{Evaluation metrics} To evaluate the performance, we employ three different standard evaluation metrics: BLUE~\cite{papineni2002bleu}, METEOR \cite{banerjee2005meteor}, and CIDEr \cite{vedantam2015cider}.

\subsection{The Effect of Different CNN Encoders}

\begin{table}[t]
	\centering
	\footnotesize
	\caption{Experiment results on the MSVD dataset. We use different features to verify our hLSTMat method. }
	\begin{tabular}{l|cccc|c}
		\hline
		Model            & B@1   & B@2   & B@3   & B@4   & METEOR  \\ \hline 
		C3D            		  & 79.9  & 68.2  & 58.3  & 47.5  & 30.5  \\ 
		GoogleNet             & 80.8  & 68.6  & 58.9  & 48.5  & 31.9  \\
		Inception-v3          & 82.7  & 72.0  & 62.5  & 51.9  & 33.5  \\ 
		ResNet-50			  & 80.9  & 69.1  & 59.5  & 49.0  & 32.3   \\
		ResNet-101            & 82.2  & 70.9  & 61.4  & 50.8  & 32.7 \\
		ResNet-152            & \textbf{82.9} & \textbf{72.2} & \textbf{63.0} & 
		\textbf{53.0} & \textbf{33.6}  \\ \hline
	\end{tabular}
	\label{tab.result_features}
\end{table}

\begin{table}[t]
	\centering
	\caption{Experimental Results on MSR-VTT dataset.  We use different features  (i.e., C3D, Inception-v3 and ResNet-152) to further verify our hLSTMat. }
	\label{tab.single_fusion_msrvtt}
	\begin{tabular}{l||c|c|c|c|c|c}
		\hline
		Model                & B@1  & B@2  & B@3  & B@4  & METEOR & CIDEr \\
		\hline
		C3D                  & 75.4 & 61.0 & 48.3 & 36.7 & 25.8   & 37.8  \\
		Inception-v3         & 72.7 & 60.3 & 48.7 & 38.2 & 26.1   & 43.1  \\
		ResNet-152              & \textbf{73.9} & \textbf{61.6} & \textbf{49.5} & \textbf{38.4} & \textbf{26.3}   & \textbf{43.4}  \\
		\hline
	\end{tabular}
\end{table}

\begin{table}[t]
	\centering
	\footnotesize
	\caption{The effect of different components and the comparison with the state-of-the-art methods on the MSVD dataset. The default encoder for all methods is ResNet-152.}
	\begin{tabular}{@{}l@{}||c|c|c|c|@{}c@{}|@{}c@{}}
		\hline
		Model   & B@1   & B@2   & B@3   & B@4   & METEOR   & CIDEr \\ \hline 
		basic LSTM            
		& 80.6     & 69.3     & 59.7     & 49.6     & 32.7 & 69.9  \\
		MP-LSTM \cite{venugopalan2014translating}          
		& 81.1     & 70.2     & 61.0     & 50.4  & 32.5  & 71.0  \\ 
		SA \cite{yao2015describing}
		& 81.6     & 70.3     & 61.6     & 51.3  & 33.3  & 72.0 \\ \hline
		basic+adaptive attention & 80.9 & 69.7 & 61.1& 50.2& 31.6&71.5\\   
		hLSTMt
		& 82.5  & 71.9  & 62.0  & 52.1  & 33.3  & 73.5   \\
		hLSTMat    & \textbf{82.9} & \textbf{72.2} & \textbf{63.0} & 
		\textbf{53.0} & \textbf{33.6} & \textbf{73.8} \\ \hline
	\end{tabular}
	\label{tab.result_methods}
\end{table}

To date, there are 6 widely used CNN encoders including C3D, GoogleNet, Inception-V3, ResNet-50, ResNet-101 and ResNet-152 to extract visual features. In this sub-experiment, we study the influence of different CNN encoders on our framework. The experiments are conducted on the MSVD and MSR-VTT datasets, and the results are shown in Tab.~\ref{tab.result_features} and Tab.~\ref{tab.single_fusion_msrvtt}. 

From Tab.~\ref{tab.result_features}, we find that by taking ResNet-152 as the visual decoder, our method performs the best with 82.9\% B@1, 72.2\% B@2, 63.0\% B@3, 53.0\% B@4 and 33.6\% METEOR, while Inception-v3 is a strong competitor, with 82.7\% B@1, 72.0\% B@2, 62.5\% B@3, 51.9\% B@4 and 33.5\% METEOR. However, the gap between ResNet-152 and Inception-v3 is very small. 
Compared with other appearance features, C3D obtains the lowest scores for both BLUE and METEOR. Unlike GoogleNet, Inception and ResNet, which can be pre-trained on a large still image classification dataset (e.g., ImageNet), the C3D is trained on video dataset \cite{tran2015learning}, and the available datasets for video action classification (e.g., Sport1M and UCF101) are still rather small. Therefore, we can conclude that in general, GoogleNet is worse than ResNet, and for ResNet, deeper networks result in a better performance. Also, the appearance features outperform the motion features for video captioning.

To further evaluate the performance of different features, we conducted more experiments on the MSR-VTT dataset by comparing C3D, Inception-v3 and ResNet-152, and show the results in Tab.~\ref{tab.single_fusion_msrvtt}. We can have similar observation to Tab.~\ref{tab.result_features} that ResNet-152 performs the best and C3D obtains the worst results.

%\vspace{-0.2cm}
\subsection{Architecture Exploration and Comparison}

In this sub-experiment, we explore the impact of three proposed components, the hierarchical LSTMs, temporal attention, and adaptive attention. Specifically, we compare the basic LSTM proposed in Sec.3.1 ``basic LSTM'',  ``basic+adaptive attention'' which adds adaptive attention to the basic LSTM, ``hLSTMt'' which removes the adaptive mechanism from the hLSTMat, and ``hLSTMat'', as well as comparing them with the state of the art methods: MP-LSTM \cite{venugopalan2014translating} and SA \cite{yao2015describing}. {In order to conduct a fair comparison, all the methods take ResNet-152 as the encoder.} We conduct the experiments on the MSVD dataset and the results are shown in Tab.~\ref{tab.result_methods}.

Tab.~\ref{tab.result_methods} shows that in general, our hLSTMat achieves the best results in all metrics with 82.9\% B@1, 72.2\% B@2, 63.0\% B@3, 53.0\% B@4, 33.6\% METEOR and 73.8\% CIDEr. 
By comparing ``basic LSTM'' with ``SA'' which adds temporal attention to the basic LSTM, we can observe that the temporal attention can improve the performance of video captioning. Specifically, the performance is improved by 1\% B@1, 1\% B@2, 1.9\% B@3, 1.7\% B@4, 0.6\% METEOR and 2.1\% CIDEr.  
Moreover, by comparing ``basic LSTM'' with ``basic+adaptive attention'', and ``hLSTMt'' with ``hLSTMt'', we find that adaptive attention plays an important role for video captioning. This indicates that the adaptive attention mechanism deciding whether to look at the visual information or language context information is beneficial for more accurate caption prediction. However, the performance improvement of adaptive attention is not as significant as temporal attention.
By comparing  ``hLSTMt'' with ``SA'', it is observed that the a significant improvement is achieved by the hierarchical LSTMs. 
The performance is improved by 0.9\% B@1, 1.6\% B@2, 0.4\% B@3, 0.8\% B@4, 0\% METEOR and 1.5\% CIDEr.   
%We also add one-layer LSTM and adaptive attention as an additional baseline. Results show that the adaptive attention mechanism can improve the performance.

\begin{table}[t]
	\centering
	\caption{The Performance of three variants architectures of temporal based hLSTMat and spatial based hLSTMat on the MSVD dataset. C, R indicate C3D and ResNet-152, respectively. T, S indicates Temporal attention and Spatial attention.}
	\label{tab.multi_fusion_msvd}
	\begin{tabular}{@{}l@{}||c|c|c|c|@{}c@{}|@{}c@{}}
		\hline
		Model                & B@1  & B@2  & B@3  & B@4  & METEOR & CIDEr                    \\
		\hline
		Motion(C)                  & 79.9 & 68.2 & 58.3 & 47.5 & 30.5   & 59.5                     \\
		Appearance(R)               & 82.9 & 72.2 & 63.0 & 53.0 & 33.6   & \textbf{73.8}                     \\
		ConF (C+R)    & 81.9 & 71.5 & 62.3 & 52.5 & 32.2   & 65.1                     \\
		ParA (C+R)  & 82.5 & 72.4 & 63.7 & 53.6 & 32.3   & \multicolumn{1}{c}{66.8} \\
		Two-stream (C+R) & \textbf{83.4} & \textbf{73.4} & \textbf{64.1} & \textbf{54.0} & \textbf{33.2}   & \multicolumn{1}{c}{71.3} \\ 
		\hline
		Temporal-Resnet        & 82.9 & 72.2 & 63.0 & 53.0 & 33.6   & \textbf{73.8}                     \\
		Spatial-Resnet         & 82.7 & 72.7 & 63.2 & 53.5 & 33.2   & 72.3                     \\
		ParA (S+T)  & 83.1 & 73.1 & 63.7 & 53.7 & \textbf{33.9}   & \multicolumn{1}{c}{69.8} \\
		Two-stream (S+T) & \textbf{83.3} & \textbf{73.6} & \textbf{64.6} & \textbf{54.3} & 33.5   & \multicolumn{1}{c}{72.8} \\ 
		\hline
		
	\end{tabular}
\end{table}

\begin{table}[t]
	\centering
	\caption{The Performance of three variants architectures of temporal based hLSTMat and spatial based hLSTMat on the MSR-VTT dataset. C, R indicate C3D and ResNet-152, respectively. T, S indicates Temporal attention and Spatial attention.}
	\label{tab.multi_fusion_msrvtt}
	\begin{tabular}{@{}l@{}||c|c|c|c|@{}c@{}|@{}c@{}}
		\hline
		Model                & B@1  & B@2  & B@3  & B@4  & METEOR & CIDEr \\
		\hline
		Motion(C)            & 75.4 & 61.0 & 48.3 & 36.7 & 25.8   & 37.8  \\
		Appearance(R)        & 73.9 & 61.6 & 49.5 & 38.4 & 26.3   & \textbf{43.4}  \\
		ConF (C+R)           & 74.3 & 61.7 & 49.1 & 37.4 & 25.9   & 41.6  \\
		ParA (C+R)           & 74.5 & 61.1 & 48.8 & 38.1 & 26.0   & 42.3  \\
		{Two-stream (C+R)} & \textbf{75.2} & \textbf{62.7} & \textbf{50.3} & \textbf{38.7} & \textbf{26.8}   & {41.9}  \\
		\hline
		Temporal-Resnet      & 73.9 & 61.6 & 49.5 & 38.4 & 26.3   & \textbf{43.4}  \\  
		Spatial-Resnet       & 74.2 & 60.9 & 48.8 & 38.3 & 26.3   & 40.8  \\
		ParA (S+T)           & 75.2 & 61.4 & 48.9 & 38.0 & 26.4   & 39.7  \\
		{Two-stream (S+T)} & \textbf{76.2} & \textbf{62.9} & \textbf{50.6} & \textbf{39.7} & \textbf{27.0}   & 42.1  \\
		\hline
	\end{tabular}
\end{table}

\begin{table}[t]
	\centering
	\caption{The performance comparison with the state-of-the-art methods on MSVD dataset. (V) denotes VGGNet, (O) denotes optical flow, (G) denotes GoogleNet, (C) denotes C3D and (R) denotes ResNet-152.}
	\begin{tabular}{@{}l@{}||c|c|c|c|@{}c@{}|@{}c@{}}
		\hline
		Model            & B@1   & B@2   & B@3   & B@4   & METEOR  & CIDEr  \\ \hline
		Random Choice    
		&  43.3    &  21.0  & 11.6    & 5.1  & 12.8    & 1.5   \\
		S2VT(V) \cite{venugopalan2015sequence}    
		&  -    &  -    &  -    & -     & 29.2    & -   \\
		S2VT(V+O)    
		&  -    &  -    &  -    & -     & 29.8    & -   \\
		HRNE(G) \cite{pan2015hierarchical}
		& 78.4  & 66.1  & 55.1  & 43.6  & 32.1    & - \\
		HRNE-SA (G)
		& 79.2  & 66.3  & 55.1  & 43.8  & 33.1    & - \\
		LSTM-E(V)\cite{pan2016jointly}
		& 74.9  & 60.9  & 50.6  & 40.2  & 29.5    & - \\
		LSTM-E(C)      & 75.7  & 62.3  & 52.0  & 41.7  & 29.9    & - \\
		LSTM-E(V+C)        
		& 78.8  & 66.0  & 55.4  & 45.3  & 31.0    & - \\
		p-RNN(V) \cite{Yu_2016_CVPR}
		& 77.3  & 64.5  & 54.6  & 44.3  & 31.1    & 62.1\\
		p-RNN(C)       & 79.7  & 67.9  & 57.9  & 47.4  & 30.3  & 53.6 \\
		p-RNN(V+C)  
		& 81.5  & 70.4  & 60.4  & 49.9  & 32.6  & 65.8 \\ 
		LSTM-TSA$_{I}$ \cite{pan2016video} 
		& 81.0  & 69.6  & 60.2  & 50.2  & 32.4  & 71.5 \\ 
		LSTM-TSA$_{V}$  
		& 82.1  & 70.7  & 61.1  & 50.5  & 32.6  & 71.7 \\
		LSTM-TSA$_{I,V}$  
		& 82.8  & 72.0  & 62.8  & 52.8  & 33.5  & \textbf{74.0} \\ \hline
		hLSTMat (T)
		& 82.9  & 72.2  & 63.0  & 53.0  & \textbf{33.6} &  73.8 \\ 
		ParA (S+T)
		& 83.1  & 73.1  & 63.7  & 53.7  & 33.9 & 69.8   \\
		\textbf{Two-stream (S+T)} 
		& \textbf{83.3} & \textbf{73.6} & \textbf{64.6} & \textbf{54.3} & 33.5   & 72.8                   \\
		
		\hline
		
	\end{tabular}
	\label{tab.result_msvd}
\end{table}

\subsection{The Effect of Spatial-temporal hLSTMat}
In this section, we aim to evaluate the performance of our proposed multiple features based hLSTMat: Two-stream and ParA, both introduced in Sec.~\ref{sec.mulLSTM}. We compare them with baselines: hLSTMat and ConF which is the simplest extension of hLSTMat. C3D (C) ResNet152 (R) are utilized in ours experiments. 
By default, we use temporal attention and ResNet152 for video captioning. 
In the first group of experiments, we use C3D (C) ResNet152 (R) as multiple features, and in the second group, we utilize temporal (T) and spatial attention (S) on ResNet152 features as multiple features.
The experiments are conducted on two datasets: MSVD and MSR-VTT, and the results are shown in  Tab.~\ref{tab.multi_fusion_msvd} and Tab.~\ref{tab.multi_fusion_msrvtt} respectively.

In the first part of Tab.~\ref{tab.multi_fusion_msvd}, we observe that in general, combining multiple features can achieve better performance than using single feature for most of the evaluation metrics of video captioning. Interestingly, using the single appearance feature achieves the best performance for CIDEr. {One possible reason is that...}. Another observation is that  Two-stream consistently outperforms ParA and ConF for all evaluation metrics. 

 The experimental results show that Two-stream performs best, ParA goes to second, and the ConF performs the worst. However, in terms of the number of training parameters, which needs to be learned, and the training time, Two-stream doubles ParA.  In Section 4.3, we find out that C3D performs worse than all the appearance features, due to lacking of enough training datasets. In other words, the potential of spatial-temporal hLSTMat networks is not fully expressed. Therefore, we conduct the third experiment by replacing C3D with Inception-v3 and the results are demonstrated in Tab.~\ref{tab.multi_fusion_msvd}. By observing the experimental results, we find out that Two-stream (I+R) performs best with all the evaluation metrics (i.e., 84.1\% B@1, 74.9\% B@2, 66.2\% B@3, 56.2\% B@4, 34.7\% METEOR and 81.1\% CIDEr). In addition, compared with ConF (C+R) and ParA (C+R), ConF (I+R) and ParA (I+R) increases the performance, respectively. Following this thought, we run spatial-temporal hLSTMat on the MSR-VTT and the results are shown in Tab.~\ref{tab.multi_fusion_msrvtt}. From which, we can see that Two-stream (I+R) achieves the best results with 77.1\% B@1, 64.0\% B@2, 51.3\% B@3, 40.0\% B@4, 27.3\% METEOR and 45.7\% CIDEr.

\begin{table}[t]
	\centering
	\caption{The performance comparison with the state-of-the-art methods on MSR-VTT dataset.}
	\begin{tabular}{l||c|c}
		\hline
		Model                  		& B@4   & METEOR   \\  \hline
		Random Choice            & 3.2 & 10.5 \\
		MP-LSTM (V)            & 34.8 & 24.8 \\
		MP-LSTM (C)            	& 35.4 & 24.8 \\
		MP-LSTM (V+C)        & 35.8 & 25.3 \\
		SA (V)       	        & 35.6 & 25.4 \\
		SA (C)                    & 36.1 & 25.7 \\
		SA (V+C)   	    	& 36.6 & 25.9 \\ \hline
		Rank-1 v2t\_navigator         & 40.8 & 28.2 \\
		Rank-2 Aalto    	& 39.8 & 28.2 \\ 
		Rank-3 VideoLAB     	& 39.1 & 27.7 \\
		Rank-4 ruc-uva     	& 38.7 & 26.9 \\
		Rank-5 Fudan-ILC     	& 38.7 & 26.8 \\ \hline
		hLSTMat (R)    		            & 38.3 & 26.3 \\ 
		ParA(S+T)                 &38.5  & 25.7 \\
		Two-stream (S+T) & \textbf{39.7} & \textbf{27.0} \\ \hline
	\end{tabular}
	\label{tab.result_msr}
\end{table}

\begin{table}[t]
	\centering
	\caption{The performance comparison with the state-of-the-art methods on LSMDC dataset.}
	\begin{tabular}{l||l|l|l|l}
		\hline
		Methods              & CIDEr          & B@4         & METEOR         & ROUGE          \\ \hline
		BUPT CIST AI lab     & 0.072          & 0.005          & \textbf{0.075} & 0.152 \\ 
		IIT Kanpur           & 0.042          & 0.004          & 0.070          & 0.138          \\ 
		Aalto University     & 0.037          & 0.002          & 0.033          & 0.069          \\ 
		Shetty and Laaksonen & 0.044          & 0.003          & 0.046          & 0.108          \\  
		S2VT                 & 0.082          & 0.007 & 0.070          & 0.149          \\ 
		Base-SAN             & 0.090          & 0.005          & 0.066          & 0.150          \\
		CT-SAN               & 0.100          & \textbf{0.008}          & 0.071          & \textbf{0.159}          \\ \hline
		hLSTMat(I)               & 0.097 & 0.007 & 0.057          & 0.150          \\ 
		hLSTMat(R)               & 0.102 & 0.007 & 0.058          & 0.147 \\
		\eat{	ParA (R+I)               & 0.086 & 0.007 & 0.058          & 0.152          \\ }
		Two-stream (R+I)               & \textbf{0.104} & 0.007 & 0.056          & 0.146 \\\hline
	\end{tabular}
	\label{tab.result_lsmdc}
\end{table}

\begin{figure*}[t]
	\centering
	\includegraphics[width=1.0\linewidth]{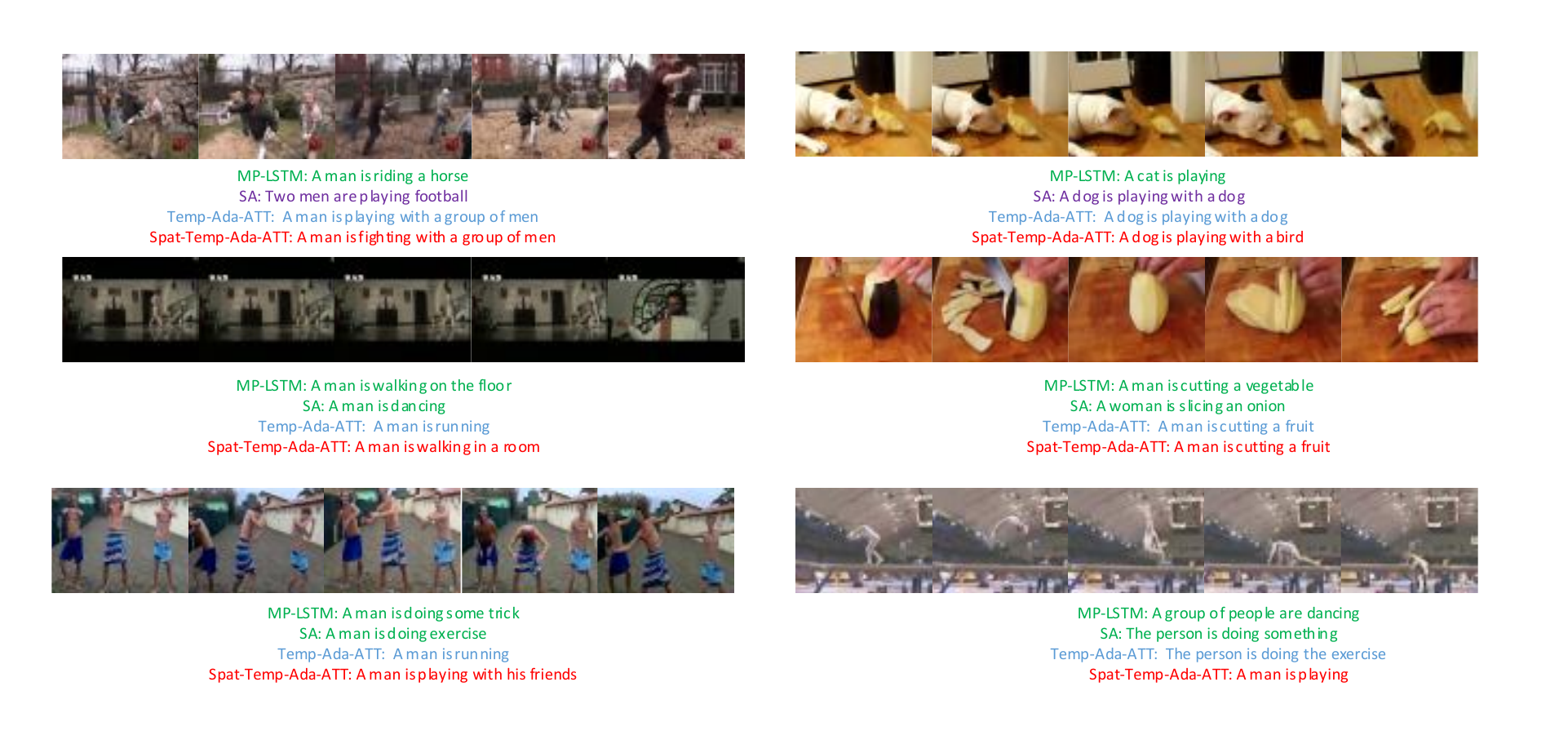}
	\caption{Examples of sentences produced by MP-LSTM and the proposed method: Temporal Adaptive attention model and Spatial Temporal Adaptive attention. Each video is represented by five frames.}
	\label{fig:qua-ana}
\end{figure*}
\begin{figure}[t]
	\centering
	\includegraphics[width=1\linewidth]{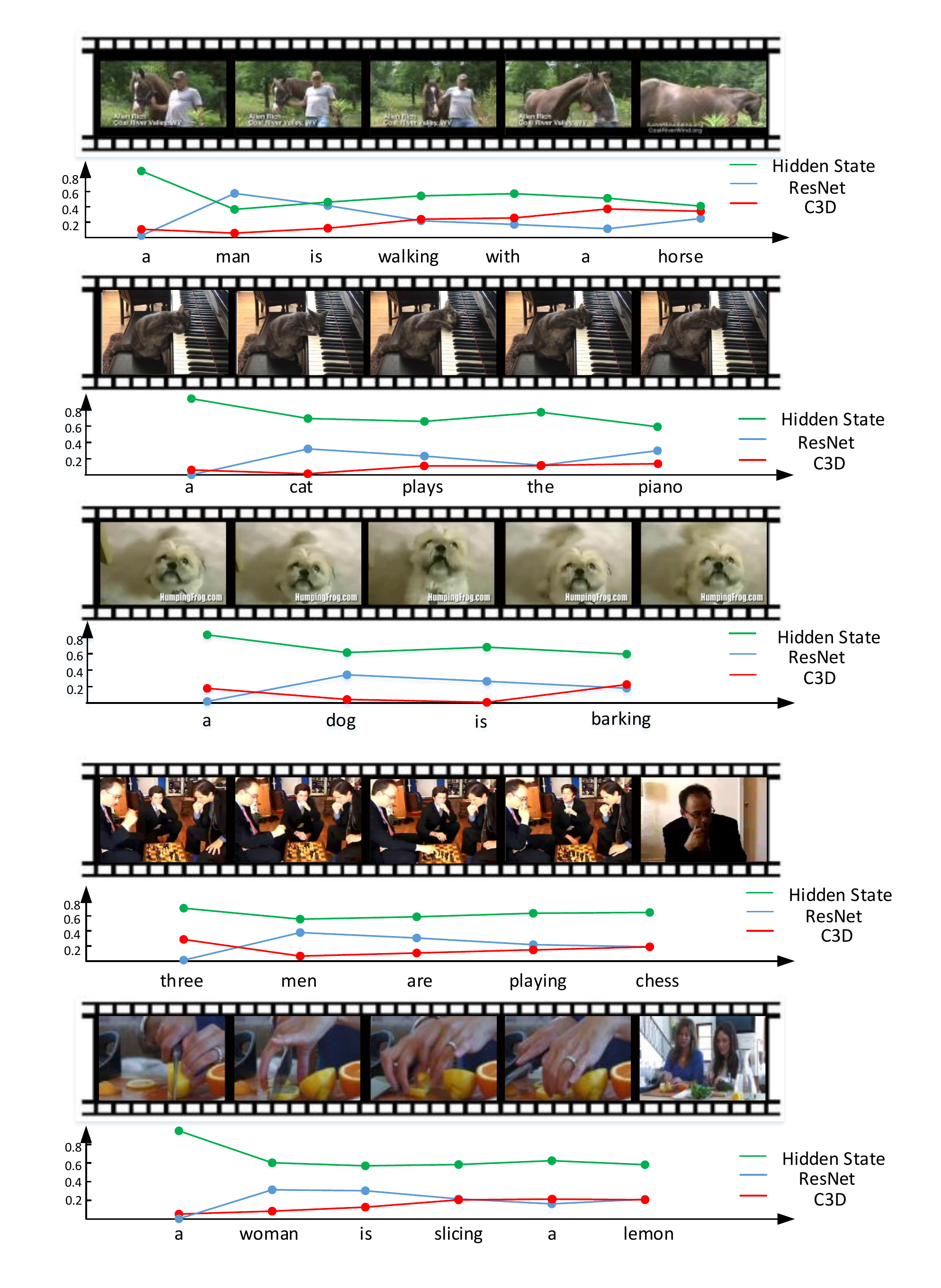}
	\caption{Examples of attention weights changes along with the generation of captions.}
	\label{fig:weight_display}
\end{figure}

\subsection{Compare with the-state-of-the-art Methods}

%To evaluate the performance of our proposed algorithm, we compare it with several state-of-the-art methods. %We also compare our hLSTMt to our hLSTMat only.

\subsubsection{Results on MSVD dataset} In this subsection, we show the comparison of our approach with the baselines on the MSVD dataset. Some of the above baselines only utilize video features generated by a single deep network, while others (i.e., S2VT, LSTM-E and p-RNN) make uses of both single network and multiple network generated features. Therefore, we first compare our method with approaches using static frame-level features extracted by a single network. In addition, we compare our spatial-temporal approaches with methods utilized two deep features. The results are shown in Tab.~\ref{tab.result_msvd} and we have the following observations:
\begin{itemize}
	\item Compared with the best counterpart (i.e., p-RNN) which only takes spatial information, our method has 8.7\% improvement on B@4 and 2.5\% on METEOR.
	\item  The hierarchical structure in HRNE reduces the length of input flow and composites multiple consecutive input at a higher level, which increases the learning capability and enables the model encode richer temporal information of multiple granularities. Our approach (53.0\% B@4, 33.6\% METEOR) performs better than HRNE (43.6\% B@4, 32.1\% METEOR) and HRNE-SA (43.8\% B@4, 33.1\% METEOR). This shows the effectiveness of our model. 
	\item Our hLSTMat (53.0\% B@4, 33.6\% METEOR) can achieve better results than our hLSTMt (52.1\% B@4, 33.3\% METEOR). This indicates that it is beneficial to incorporate the adaptive temporal attention into our framework.
	\item Our hLSTMat achieves a good performance (53.0\% B@4 and 33.6\% METEOR) using static frame-level features compared with approaches combining multiple deep features. For S2VT (V+O), LSTM-E (V+C) and p-RNN (V+C), they use two networks VGGNet/GoogleNet and optical flow/C3D to capture videos' spatial and temporal information, respectively. Compared with them, our hLSTMat only utilizes ResNet-152 to capture frame-level features, which proves the effectiveness of our hierarchical LSTM with adaptive temporal attention model.  
\end{itemize}

In addition, simultaneously considering both spatial and temporal video information can enhance the video caption performance. VGGNet and GoogleNet are used to extract spatial information, while optical flow and C3D are utilized for capturing temporal information. For example, compared with LSTM-E (V) and LSTM-E (C), LSTM-E (V+C) achieves higher B@4 (45.3\%) and METEOR (31.0\%). For p-RNN, p-RNN (V+C) (49.9\% B@4 and 32.6\% METEOR ) performs better than both p-RNN(V) (44.3\% B@4 and 31.3\% METEOR) and p-RNN(C) (47.4\% B@4 and 30.3\% METEOR). Compared with all two features based approaches LSTM-E (V+C), p-RNN (V+C) and LSTM-TSA$_{I,V}$, our Two-stream (R+I) achieves the best performance with 84.1\% B@1, 74.9\% B@2, 66.2\% B@3, 56.2\% B@4, 34.7\% METEOR and 81.1\% CIDEr.  

We adopt questionnaires collected from ten users with different academic backgrounds. Given a video caption, users are asked to score the following aspects: 1) Caption Accuracy, 2) Caption Information Coverage, 3) Overall Quality. Results show that our method outperforms others at 'Overall Quality', and 'Caption Accuracy' with small margin. But it has lower value for 'Information coverage' than p-RNN.

\subsubsection{Results on MSR-VTT dataset} 
We compare our model with the state-of-the-art methods on the MSR-VTT dataset, and the results are shown in Tab.~\ref{tab.result_msr}. Our Two-stream model performs the best on all metrics with 40.0\% @B4 and 27.3\% METEOR, while our hLSTMat (R) goes to second with 38.3\% @B4 and 26.3\% METEOR. Compared with our method  hLSTMat using only temporal attention, the performance is improved by 1.7\% for @B4, and 1.0\% for METEOR. This verifies the effectiveness of our hLSTMat approach. In addition, our two-stream (R+I) increase the state-of-the-art method SA (V+C) by 3.4\% on B@4 and 1.4\% METEOR. This proofs the importance of two-stream networks. Besides, We compared our method with the MSR-VTT 2016 challenge results and we can see that our method can attain comparable performance. The leaderboard of MSR-VTT 2016 challenge can be easily found on the Internet.
\footnote{http:http://ms-multimedia-challenge.com/2016/leaderboard}

% 37.4\% @B4 and 26.1\% METEOR using our temporal attention model and  adding our adjusted mechanism.

\subsubsection{Results on LSMDC dataset} 
In this section, we report the performance of different methods on the LSMDC dataset, see Tab. \ref{tab.result_lsmdc}. All the previous methods are LSMDC participants, which have no obligation to disclose their identities or used technique. Among comparable methods, our approach Two-stream obtains the highest CIDEr (10.4\%). In terms of B@4 and ROUGE, CT-SAN ranks first with 0.8\% and 15.9\%, respectively. And BUPT CIST AI lab achieves best performance on METEOR. In addition, compared with MSR-VTT (i.e., each video contains approximately 20 natural language sentences), the LSMDC dataset does not contain sufficient descriptions for each video.  For 118,081 video clips, it only has 118,114 sentences. In average, each video only has one language description. However, in reality, human can provides multiple natural language sentences for describing a video. In other words, a video can be described from different aspects, but the the ground truth only provides one description. This might be the major reason why all the methods gain low performance scores on this dataset in terms of all evaluation metrics. 

\subsubsection{Qualitative analysis} 

In this section, we display several examples of actual videos in different methods. Here, we use three model MP-LSTM, our proposed Temporal Adaptive attention and Spatial-Temporal Adaptive attention to generate sentence and their sentences in green, blue and read respectively. Compared with

\subsection{Feature Weights Analysis}
In this section, we discuss the feature weights learned by the spatial-temporal framework: ParA by demonstrating some examples in Fig.~\ref{fig:weight_display}. More specifically, we aim to answer how ParA leverages the three important factors including the motion feature (C3D), appearance feature (ResNet152) and language context information (i.e., hidden state value), and what are the their weight trends in a captioning generation process? 

Fig.~\ref{fig:weight_display} provides five video captioning examples and each shows the weights change of ResNet152 (R), C3D (C) and hidden state (H). The generated natural language sentences are shown under the horizontal time axis, and each word is positioned at the time step when it is generated. The vertical axis stands for the weight scores and their sum weights equals to 1. From the five examples, we can see that the syntactic information is quite essential to captioning and the weight of language content information is always larger than other two feature weights. This is because the H not only contains the syntactic information but also contains the previous appearance and motion information.   

If we look at the trend over time, we can see that 1) when generating non-visual words such as ``a'', ``the'',``are'' and ``is'', the over all trend of H is upwards. This indicates that context information plays a more important role in non-visual words generation. In the second example, weights of H is quite high (i.e., both more than 0.7) for generating ``a" and ``the". 2) when generating the subjective in a sentence, such as ``man'', ``cat'',``dog'', ``men'' or ``women'', the trend of R grows dramatically, while the trend of C goes down.  3) when generating actions such as ``plays the piano", ``playing chess" and ``slicing a lemon", the trend of C is upwards. In general, by comparing appearance and motion features, an interesting phenomenon is found: appearance feature's weight tends to be larger in generating nouns such as ``man", ``cat", ``dog" and ``horse", while motion feature's weight tends to be higher in generating verbs such as ``walking", ``plays", ``barking" and``slicing.

\section{hLSTMat for Image Captioning}
\label{sec.imageCap}
\begin{figure*}[t]
	\centering
	\includegraphics[width=1\linewidth]{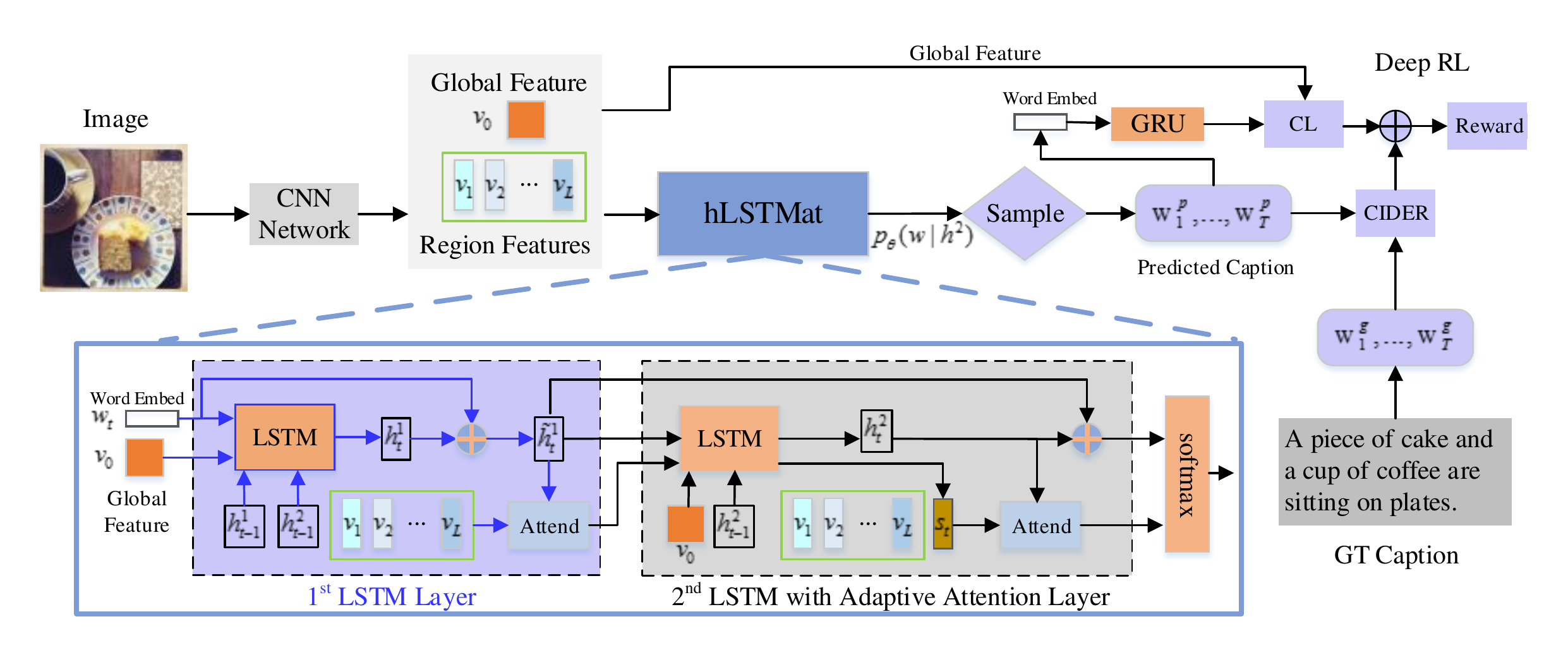}
	\caption{The proposed framework for DA, our model uses residual shortcut connection to improve information flow through two LSTMs. And adaptive attention is applied to calculate weights of features when predicting new word.}
	\label{fig.framework2}
\end{figure*}

\eat{In this section, we instantiate hLSTMat and apply it to the task of image captioning. The overall architecture of the DA net is shown in Fig.~\ref{fig.framework2}. 
%To ease understanding, our full framework is described in parts. 
We begin by introducing the image encoding. Then, we describe deliberate attention mechanism and reinforcement module.
%In particular, our deliberate attention model is designed by extending traditional soft attention model for caption generation, while the reinforcement learning component is applied to guide the training process of the deliberate attention module.
}

In this subsection, we instantiate our hLSTMat and apply it to the task of image captioning. As shown in Fig.~\ref{fig.framework2}, there are three major components: 1) a CNN Encoder, 2) an attention based hierarchical LSTM decoder and 3) the losses. We give the details for each component below.

\begin{figure}[t]
	\centering
	\scriptsize
	\includegraphics[width=1\linewidth]{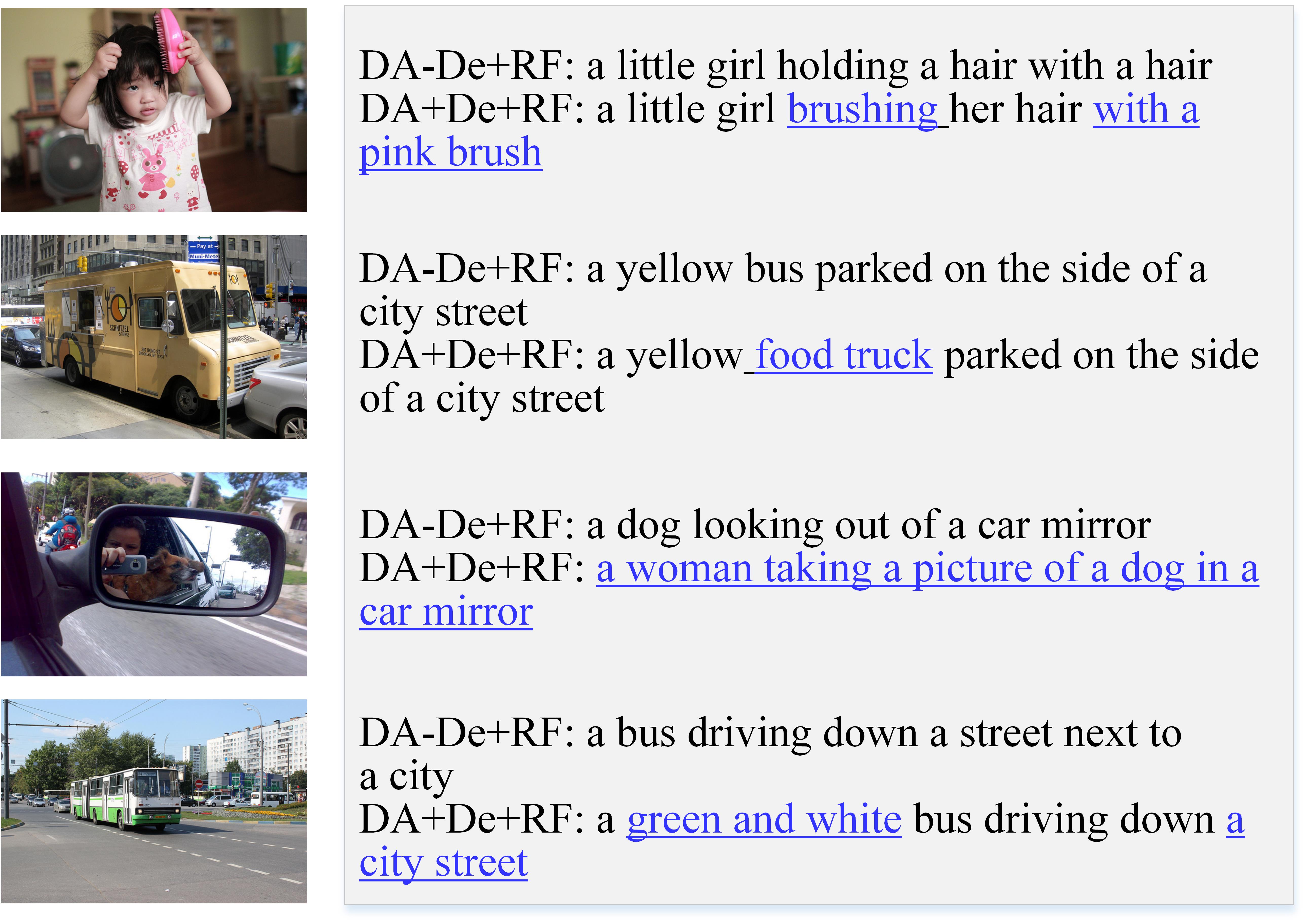}
	\caption{Examples of image captions generated by our DA. For each image, the first and second sentences are generated by the first-pass (DA-De+RF) and the second-pass respectively (DA+De+RF). This demonstrates that our deliberate attention can generate more precise concepts (e.g., ``food truck'' vs ``bus'') and more reasonable descriptions (e.g., ``a women taking a picture of a dog in a car mirror'' vs ``a dog looking out of a car mirror'') by using a deliberate process.}
	\label{fig.firstLayer}
\end{figure}

\subsection{CNN Encoder}
%Image encoder is an essential part of image captioning and it is used to extract visual content information of images. CNN designed for image classification is usually adopted to extract a global visual image feature, while R-CNN designed for object detection is usually employed to derive region visual features. Compared with a CNN-based visual feature which is obtained within a global context, a R-CNN based region visual feature contains rich information about a particular object. 
In this paper, we adopt a pretrained ResNet-101 \cite{Resnet} (pool5) to extract global visual feature, and use Faster R-CNN \cite{fasterrcnn} to produce bounding boxes and then apply them on the pretrained ResNet-101 to extract $L$ region features. For simplicity, given an image $\textbf{x}$, we extract $L+1$ image features, represented as $\left\{{\textbf{v}_g},\textbf{V}\right\}$, where ${\textbf{v}_g}$ is the global visual feature and $\textbf{V} = \left\{ {{\textbf{v}_1}, \cdots ,{\textbf{v}_L}} \right\}$ indicates the $L$ region visual features.

\subsection{Hierarchical LSTM with Adaptive Attention} 
Our hLSTMat consists of two LSTM blocks. 
The first LSTM aims to prepare the hidden states and visual attention for generating a preliminary version of the captions, while the second LSTM is applied as a proofreading process to refine them. Both LSTM are built upon the basic LSTM (Eq.~\ref{eq.lstm}).

For the $t$-th time step, ${\textbf{y}_t}$ is the input vector of the LSTM unit, ${\textbf{h}_t}$ is the output vector of the LSTM unit, and ${\textbf{h}_{t - 1}}$ is the output of the LSTM unit at the $t$-$1$ time step. For simplicity, we refer to the operation procedure of basic LSTM with the following notation:
\begin{equation}
{\textbf{h}_t} = LSTM({\textbf{y}_t},{\textbf{h}_{t - 1}})
\end{equation}

\subsubsection{The First LSTM}
We modify the basic LSTM to generate an initial text sequence feature as depicted in Fig.~\ref{fig.framework2}. We define the input of the LSTM unit as:
\begin{equation}
\textbf{y}_t^1 = [{\textbf{v}_g},\textbf{h}_{t - 1}^2,{\textbf{w}_t}]
\end{equation}
where ${\textbf{v}_g}$ is the global image visual feature, $\textbf{h}_{t - 1}^2$ is the output of the second LSTM layer, at the $t$-$1$ time step, and ${\textbf{w}_t}$ is the feature of the current word derived by embedding an one-hot word vector. 
It is easy to understand that the current hidden states are based on the global visual feature, the previous hidden states and the $t$-th word. We further use $\textbf{h}_{t - 1}^2$ from a higher-level LSTM in order to utilize the more precise information to guide the learning of $\textbf{h}_t^1$. The we get ${\textbf{h}_t^1} = LSTM({\textbf{y}_t^1},{\textbf{h}^1_{t - 1}})$.
%When we obtained the $x_t^1$, we apply the basic LSTM since it might be able to connect previous information to the present prediction task. Specifically, we believe that visual global feature, previous output and the $t$-th high frequency word might inform the generation of the $t$-th word.

Traditionally, the hidden state of the LSTM is directly applied to guide which region should be focused on. The LSTM provides a temporal shortcut path to avoid vanishing gradients. Here, we provide an additional word shortcut from the $t$-th high frequency word for efficient training. As a result, we use a residual shortcut connection to further reduce vanishing gradients:
\begin{equation}
\tilde {\textbf{h}}_t^1 = {\textbf{W}_{rd}}[{\textbf{w}_t};\textbf{h}_t^1]	
\end{equation}
where $\textbf{h}_t^1$ is the hidden states at the $t$-th step, $\textbf{W}_{rd}$ is the parameter to be learned, and [;] is a concatenation operation. 

Given $L$ image region visual features $\left\{ {{\textbf{v}_1}, \cdots ,{\textbf{v}_L}} \right\}$ and the hidden state  $\tilde {\textbf{h}}_t^1$, we aim to selectively utilize certain region visual features by defining a visual attention mechanism below:
\begin{align}
\textbf{e}_t^1 &= \textbf{w}_{e1}^T\tanh ({\textbf{W}_{v1}}\textbf{V},{\textbf{W}_{h1}}\tilde h_t^1)\\
\alpha _t^1 &= soft\max (\textbf{e}_t^1)
\end{align}	
where $\textbf{V} = [{\textbf{v}_1},{\textbf{v}_2}...,{\textbf{v}_L}]$ indicates the $L$ image region features. And $\textbf{w}_{e1}^T,{\textbf{W}_{v1}},{\textbf{W}_{h1}}$ are parameters to be learned. $\alpha _t^1 \in {\mathbb{R}^L}$ is the attention weight, which is then applied on region features to locate the important visual information:
\begin{equation}
\hat{\textbf{v}}_t^1 = \sum\nolimits_{l = 1}^L {\alpha _{l,t}^1{\textbf{v}_l}} 
\end{equation}
where $\hat{\textbf{v}}_t^1$ is the attended visual information which can be used together with $\tilde {\textbf{h}}_t^1$ to generate the primary $t$-th word.

\subsubsection{The Second LSTM with Adaptive Attention}
By integrating the softmax layer and the loss functions to the first LSTM layer, we can generate a preliminary word at each step. In this subsection, we design a second LSTM layer with adaptive attention as a deliberate process to further purify the captioning results. Similar to Eq.~\ref{eq.ct}, we first define the language context information $s_t$ as:
\begin{align}
{\textbf{g}_t} &= \sigma ({\textbf{W}_x}\textbf{y}_t^2 + {\textbf{W}_h}\textbf{h}_{t - 1}^2)\\
{\textbf{s}_t} &= {\textbf{g}_t} \odot \tanh (\textbf{m}_t^2)
\end{align}
where ${\textbf{W}_x},{\textbf{W}_h}$ are parameters to be learned, $\textbf{y}_t^2$ is the input of the LSTM unit. $\odot$ is an element-wise product and $\sigma$ represents the logistic sigmoid activation. More specifically,
\begin{equation}
\textbf{y}_t^2 = [{\textbf{v}_g},\tilde {\textbf{h}}_t^1,\hat {\textbf{v}}_t^1]
\end{equation}
After we get ${\textbf{h}_t^2} = LSTM({\textbf{y}_t^2},{\textbf{h}^2_{t - 1}})$ and ${\textbf{s}_t}$ from the second LSTM, we compute an attention vector to determine when and where to look at the visual and context information. To compute the attention vector, we firstly obtain $\textbf{e}_t^2$ by: 
\begin{equation}
\textbf{e}_t^2 = \textbf{w}_{z2}^T\tanh ({\textbf{W}_{v2}}V,{\textbf{W}_{h2}}\textbf{h}_t^2)
\end{equation}
where ${\textbf{w}_{z2}^T},\textbf{W}_{v2},{\textbf{W}_{h2}}$ are parameters to be learned. $\textbf{h}_t^2$ is the LSTM output at the $t$-th time step. Next, we use the following function to calculate the attention weights $\alpha _t^2$.
\begin{equation}
\alpha _t^2 = soft\max ([\textbf{e}_t^2;\textbf{w}_{a}^T\tanh ({\textbf{W}_s}{\textbf{s}_t} + {\textbf{W}_{h3}}\textbf{h}_t^2)])
\end{equation}
where ${\textbf{w}_{a}^T},\textbf{W}_{s},{\textbf{W}_{h3}}$ are parameters to be learned. $\alpha _t^2 \in {\mathbb{R}^{L + 1}}$ contains weights for image region features and the sequential context information $\textbf{s}_t$. Finally, the attended results can be obtained by:
\begin{equation}
\hat {\textbf{v}}_t^2 = \sum\nolimits_{i = 1}^{L + 1} {\alpha _{i,t}^2{\textbf{v}_i}} 
\end{equation}
where ${{\textbf{v}_{L + 1}}}$ equals to ${\textbf{s}_t}$.

To generate the $t$-th word, we combine the output of the first layer $\tilde {\textbf{h}}_t^1$, the output of the second LSTM $\textbf{h}_t^2$ and the attended visual features $\textbf{v}_t^2$ together by the function below:
\begin{equation}
\tilde {\textbf{h}}_t^2 = {\textbf{W}_{sd}}[\tilde {\textbf{h}}_t^1;{\textbf{h}}_t^2;\hat {\textbf{h}}_t^2]	
\end{equation}
where ${\textbf{W}_{sd}}$ is the parameter to be learned. We use softmax function to calculate the probability of the $t$-th word:
\begin{equation}
{\textbf{p}_t} = softmax(\tilde {\textbf{h}}_t^2)
\label{eq.predict}
\end{equation}

\subsection{Two-step Training} 
In essential, our training process can be divided into two stages. Firstly, the parameters $\Theta$ in image captioning model are pre-trained by minimizing MLE loss (Eq.~\ref{eq:mle}), and then we fine-tune the model with reinforcement learning.

\subsubsection{Training with Reinforcement Learning} 
Inspired by the previous work \cite{discriminability,self_critical}, we consider our model introduced above as ``agent'' to interact with external environment (i.e., words, global and region visual features),  and $(\Theta )$ as the policy to conduct an action to predict a word. After the whole caption is generated, the agent observers a reward. Since CIDEr is proposed to evaluate the quality of image captioning model. We design our reward functions by combing contrastive loss (CL) with CIDEr. Next, we introduce the definition of CL.

\textbf{Contrastive Loss (CL).} Given an image ${\textbf{x}}$ and caption ${\textbf{c}}$, we obtain caption and image features by RNN and CNN, respectively. We take global image feature ${\textbf{v}_0}$ as image features. Each word in ${\textbf{c}}$ is embedded and then input into a RNN network to derive a caption feature. We define caption feature as ${\textbf{c}_0}$. Next, we map two features into a common space by $\textbf{W}_v^T$ and $\textbf{W}_c^T$, respectively:
\begin{equation}
f(\textbf{x}) = \textbf{W}_v^T{\textbf{v}_0} 
\end{equation}   
\begin{equation}
f(\textbf{c}) = \textbf{W}_c^T{\textbf{c}_0}
\end{equation}
Furthermore, cosine similarity is used to compute similarity between an image and caption:
\begin{equation}
s(\textbf{x},\textbf{c}) = \frac{{f(\textbf{x}) \cdot f(\textbf{c})}}{{\left\| {f(\textbf{x})} \right\|\left\| {f(\textbf{c})} \right\|}}
\end{equation}
Parameters in the such model are learned by minimizing the contrastive loss (CL), which is a sum of two hinge losses:
\begin{equation}
\begin{array}{l}
{\ell_{CON}}(\textbf{x},\textbf{c}) = \mathop {\max }\limits_{\textbf{c}'} {\left[ {\alpha  + {\rm{s(\textbf{x},\textbf{c}') - s(\textbf{x},\textbf{c})}}} \right]_ + }\\
\ \ \ \ \ \ \ \ \ \ \ \ \ \ \ \ \ \ + \mathop {\max }\limits_{\textbf{x}'} {\left[ {\alpha  + {\rm{s(\textbf{x}',\textbf{c}) - s(\textbf{x},\textbf{c})}}} \right]_ + }
\end{array}
\label{cl}
\end{equation}
where ${\left[ \textbf{x} \right]_ + } \equiv \max (\textbf{x},0)$. In Eq.~\ref{cl}, $(\textbf{c},\textbf{x})$ indicates that a pair of caption and image is matched. $\textbf{c}$ correctly describes image $I$. Both $(\textbf{x},\textbf{c}')$ and $(\textbf{x}',\textbf{c})$ are mismatched. $((\textbf{x},\textbf{c}'))$ suggests that $\textbf{c}'$ is the incorrect description of $\textbf{x}$, while $(\textbf{x}',\textbf{c})$ suggests that $\textbf{c}$ is the incorrect description of $\textbf{x}'$. $\alpha$ is used to ensure minimum gap between scores of $(\textbf{c},\textbf{x})$ with $(\textbf{x},\textbf{c}')$ and $(\textbf{x}',\textbf{c})$.

\textbf{CIDEr+CL} In order to optimize the parameters in our model, the objective becomes to maximize the reward obtained from reward function by learning parameters. An update is implemented by computing the gradient of the expected reward: 
\begin{equation}
{\nabla _\Theta }E[R(\hat {\textbf{c}},\textbf{x})] \approx (R(\hat {\textbf{c}},\textbf{x}) - R({{\textbf{c}}^*},\textbf{x})){\nabla _\Theta }\log p(\hat {\textbf{c}}|\textbf{x};\Theta )
\end{equation}
where $R(\hat {\textbf{c}},\textbf{x}) = CIDEr(\hat {\textbf{c}}) - {\ell_{CON}}(\hat {\textbf{c}},\textbf{x})$ is reward function. $\hat {\textbf{c}}$ is caption generated from \eqref{eq.predict}. The baseline is computed by ${\textbf{c}^*} = (BOS,\textbf{w}_1^*,...,\textbf{w}_T^*)$, which is obtained as:
\begin{equation}
\textbf{w}_t^p = \mathop {\arg \max }\limits_\textbf{w} (\textbf{w}|\textbf{w}_{0,...t - 1}^p,\textbf{x})	
\end{equation}

\section{Experiments for Image Captioning}
\label{sec.exp2}

%Firstly, we study the influence of the parameters in our algorithm. Then, we compare our results with state-of-the-art algorithms on three standard datasets.
%\footnote{The code for SHPL will be released for public use.}

\subsection{Datasets}
In this paper, we utilize two datasets including COCO~\cite{COCO} and Flickr30K~\cite{Flickr30K} to evaluate the performance of our proposed DA network. 

\textbf{COCO.} It is the largest dataset for image captioning, which consists of $82,783$, $40504$ and $40,775$ images for training, validation and testing, respectively. \eat{Generally, the validation set is divided into two parts: one for off-line testing and another for validation of hyper-parameters. }In COCO dataset, each image has 5 captions annotated by human beings. Following previous work \cite{att_fcn,adaptive,discriminability,bottomup}, we use the ``Karpathy'' splits proposed in ~\cite{Karpathysplits}. In this split, $113,287$, $5,000$ and $5,000$ images are used for training, validation and testing, respectively.

\textbf{Flickr30K.} It consists of $31,783$ images collected from Flickr. In particular, each image is associated with 5 crowd-scoured descriptions. Most of the images are about human beings performing activities. Following previous work \cite{adaptive}, we use 29k images for training, 1k for validation and 1k for testing. 

For both COCO and Flickr30K dataset, we conduct a pre-processing procedure by firstly truncating captions longer than 16 words. Next, all modified captions are converted to lower case. For each dataset, we build a vocabulary. For COCO, the vocabulary contains $9,487$ words, while for Flickr30K the vocabulary has 7k words.

\subsection{Evaluation Metric}
Five generally used evaluation metrics are adopted to evaluate the performance of image captioning, including BLUE~\cite{bleu}, ROUGEL~\cite{rouge}, METEOR~\cite{meteor}, CIDEr~\cite{cider} and SPICE~\cite{spice}. More specifically, for the COCO dataset, we also report the SPICE subclass scores on 5k validation sets, including Color, Attribute, Cardinality, Object, Relation and Size. All the SPICE subclass scores are scaled up by 100.

\begin{table*}[]
	\centering
	\caption{Ablation study results obtained from the COCO dataset.}
	\label{tab.ablation}
	\begin{tabular}{l|l|l|l|l|l|l|l|l|l|l|l|l}
		\hline
		model                           & BLUE1   & BLUE4   & METEOR    & ROUGE   & CIDEr     & SPICE    & Att & Card & Col & Obj  & Rel & Size \\ \hline
		DA-De-RF                  & 74.2 & 33.7 & 26.4 & 54.6 & 104.9 & 19.4 & 9.3  & 2.1  & 11.0  & 35.5 & 5.2 & 3.8  \\
		DA-De+RF        & 78.4 & 35.2 & 27.3 & 56.5 & 117.3 & 21.1 & 9.2  & 13.0 & 10.5  & 39.2 & 5.6 & 2.8 \\ \hline
		DA+De-RF         & 75.8 & 35.7 & 27.4 & 56.2 & 111.9 & 20.5 & 10.8 & 6.1  & 14.6  & 36.8 & 5.5 & \textbf{5.6}  \\
		DA+De+RF                & \textbf{79.9} & \textbf{37.5} & \textbf{28.5} & \textbf{58.2} & \textbf{125.6} & \textbf{22.3} & \textbf{11.2} & \textbf{14.4} & \textbf{15.2}  & \textbf{40.3} & \textbf{6.4} & 3.7  \\ \hline 
	\end{tabular}
\end{table*}

\begin{figure*}[t]
	\centering
	\scriptsize
	\includegraphics[width=0.8\linewidth]{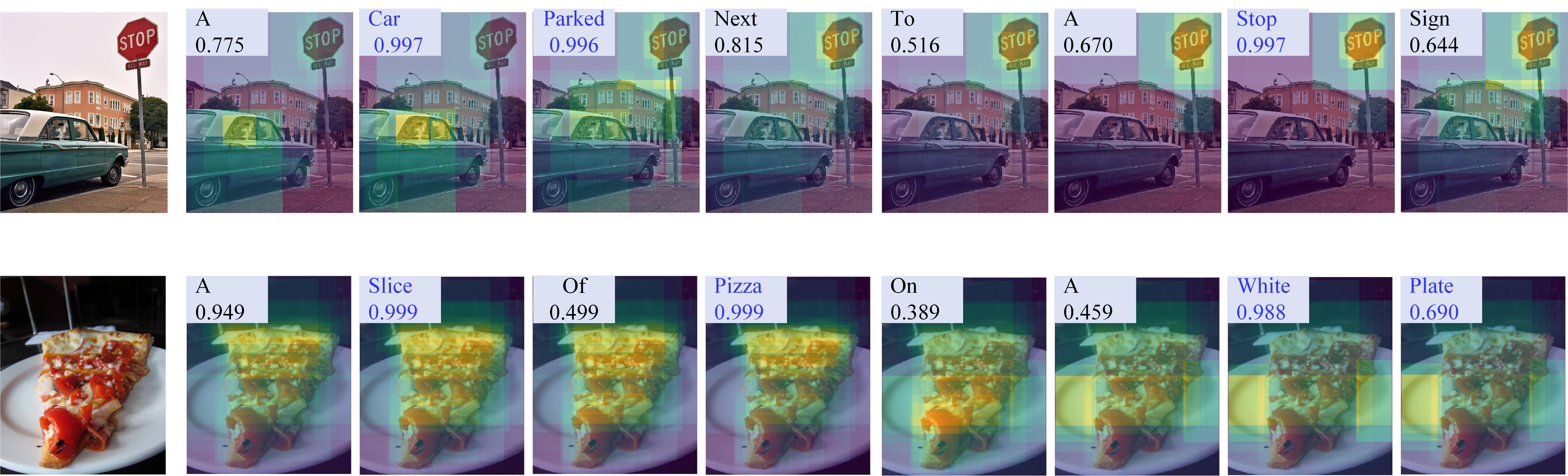}
	\caption{Visualization of first residual attention map of the DA+De+RF model. The sentence is generated by the DA+De+RF. The region with the maximum attention weight is in orange. We also show each word with the corresponding visual weight in the second residual attention block.}
	\label{fig.attention}
\end{figure*}

\subsection{Implementation Details}
To extract the global visual feature, we use pre-trained ResNet-101 \cite{Resnet} to process the input image and the output of ``pool5'' (2048-D) is used as global appearance feature. In terms of region visual features, we utilize the same region features used in \cite{bottomup}. \eat{Anderson \textit{et al.} \cite{bottomup} incorporate Faster-RCNN ~\cite{fasterrcnn} with ResNet-101 \cite{Resnet} to extract region features. The region feature extraction model is trained on ImageNet~\cite{imagenet} and Genome data~\cite{Genome}.} As a result, for each image, we obtain 36 region features and the dimension for each region feature is also 2048-D. In addition, the hidden state size of two LSTM in our DA network is set to be 512. 

To compute the contrastive loss, we use RNN as the text encoder to extract caption features and use pool5 of ResNet-101 to extract image features. The dimension of word embedding is 512, the RNN hidden state size is set as 1024 and the dimension of embedded image feature is 1024. To train the model for calculate contrastive loss, we set epochs as 27 for both COCO and Flicker30K.

\begin{table*}[]
	\centering
	\caption{Performance on Flicker30k test split. DA refers to the DA+De+RF mode in Tab.~\ref{tab.ablation}.}
	\label{tab.flickr}
	\begin{tabular}{l|l|l|l|l|l|l}
		\hline
		model          & BLEU-1   & BLEU-2   & BLEU-3   & BLEU-4   & METEOR    & CIDEr    \\ \hline
		DeepVS~\cite{deepvs}          & 57.3 & 36.9 & 24.0 & 15.7 & 15.3 & 24.7 \\
		Hard-Attention~\cite{hard_attention} & 66.9 & 43.9 & 29.6 & 19.9 & 18.5 & -    \\
		ATT-FCN~\cite{att_fcn}        & 64.7 & 46.0 & 32.4 & 23.0 & 18.9 & -    \\
		Adaptive-Attention~\cite{adaptive}      & 67.7 & 49.4 & 35.4 & 25.1 & 20.4 & 53.1 \\ \hline
		DA              & \textbf{73.8} & \textbf{55.1} & \textbf{40.3} & \textbf{29.4} & \textbf{23.0} & \textbf{66.6} \\ \hline
		Relative Improvement &6.1& 5.7& 4.9 &4.3&2.6& 13.5\\ \hline
	\end{tabular}
\end{table*}

\begin{table*}[t]
	\centering
	\caption{Single-model image captioning performance on the COCO Karpathy test split. B-4, M, R, C and S are BLUE4, METEOR, ROUGE, CIDEr and SPICE scores, respectively. All methods are based on the reinforcement learning.}
	\label{tab.coco}
	\begin{tabular}{l|l|l|l|l|l|l|l|l|l|l|l|l}
		\hline
		model                                & B4   & MR    & R   & C   & S    & Att & Card & Col & Obj  & Rel & Size \\ \hline
		SCST:Att2in~\cite{self_critical}     & 31.3 & 26.0 & 54.3 & 101.3 & -    & -    & -    & -     & -    & -   & -    \\
		SCST:Att2all~\cite{self_critical}    & 30.0 & 25.9 & 53.4 & 99.4  & -    & -    & -    & -     & -    & -   & -    \\
		ATTN+C+D(1)~\cite{discriminability}  & 36.3 & 27.3 & 57.1 & 114.1 & 21.1 & 9.5  & 10.5 & 9.3   & 39.0 & 5.9 & 2.6  \\
		Up-Down~\cite{bottomup}              & 36.3 & 27.7 & 56.9 & 120.1 & 21.4 & 10.0 & \textbf{18.4} & 11.4  & 39.1 & \textbf{6.5} & 3.2  \\ \hline
		DA                     & \textbf{37.5} & \textbf{28.5} & \textbf{58.2} & \textbf{125.6} & \textbf{22.3} & \textbf{11.2} & 14.4 & \textbf{15.2}  & \textbf{40.3} & 6.4 & \textbf{3.7}  \\ \hline
	\end{tabular}
\end{table*}

%\textbf{Training Details.} 
For training process, MLE is firstly used to pre-train the DA model. Next, we train it with reinforcement learning by using CIDEr and CL as reward value. For MLE training, the epoch is set as 150 for both COCO and Flickr30K, while for reinforcement learning the epoch is set as 200 for COCO and 150 for Flickr30K. All models are trained by using Adam and the batch size is set as 128. We initialize the learning rate with 5e-4 and update it by a decreasing factor 0.8 in every 15 epochs. When conduct testing, beam search is applied to predict captions, with beam size setting as 5.

\subsection{Ablation Study}
In order to figure out the contribution of each component, we conduct the following ablation studies on the COCO dataset. 
Specifically, we remove the deliberation process (De) and reinforcement learning (RF) respectively from our DA model, and have four experiments: DA-De-RF, DA-De+RF, DA+De-RF and DA+De+RF.  The experimental results are shown in Tab.~\ref{tab.ablation}.
%In order to test the impact of deliberation process and reinforcement learning in image captioning, we conduct four experiments: 1) without deliberation (i.e., the second residual based attention layer) and reinforcement learning, namely DA-De-RF; 2) without deliberation but with reinforcement learning, namely DA-De+RF. For both models, they use $\hat v_t^1$ and $\tilde h_t^1$ to predict the words. 3) with deliberation but without reinforcement learning, namely DA+De-RF. 4) with deliberation and with reinforcement learning, namely DA+De+RF, which is our final model. Here, we report six metrics including BLEU-1, BLEU-4, METEOR, ROUGE, METEOR, CIDEr and SPICE as well as other six SPICE subclass scores on Attribute, Cardinality, Color, Object, Relation and Size. The scores are calculated with COCO captioning evaluation tool. The experimental results are shown in Tab.~\ref{tab.ablation}

\textbf{The Influence of Deliberation.} From Tab.~\ref{tab.ablation}, we can see that with or without reinforcement learning, our DA+De models, including DA+De+RF and DA+De-RF, perform better than DA-De models (DA-De-RF and DA-De+DF). More specifically, DA+De-RF outperforms DA-De-RF on all 12 metrics, in particular with an increase of 2\% on BLUE4, 1\% on METEOR, 1.6\% ROUGE, 7\% CIDEr and 1.1\% SPICE. In addition, compared with DA-De+RF, DA+De+RF obtains higher performance on all 12 metrics. This verifies the importance of our deliberation mechanism.

In order to further demonstrate the role of deliberation mechanism, we show four visual examples in Fig.~\ref{fig.firstLayer}. In Fig.~\ref{fig.firstLayer}, each image contains two descriptions. The first sentence is generated by the DA-De+RF and the second sentence is generated by the DA+De+RF model. We can see that without deliberation process, the model can generate a sentence which contains error information (e.g., ``a hair with a hair'') or inconsistent semantic information, e.g., ``a dog looking out of a car mirror''.  With the deliberation component, our DA model can provide more precise description, such as ``a yellow food truck'' instead of ``a yellow bus''; and a more reasonable description: ``a women taking a picture of a dog in a car mirror'' instead of ``a dog looking out of a car mirror''. These examples also show that the first residual attention based layer can detect primary objects (e.g., girl, hair) and activities (e.g., holding), while the second residual attention based layer can refine the activities (e.g., brushing) and detect the relationship between objects (e.g., ``a women taking a picture of a dog'' and ``a dog in a car mirror''). %The visualization examples demonstrates the positive role of our deliberation process. 

In addition, Fig.~\ref{fig.attention} shows the attended image regions of the first residual based attention of the DA+De+RF. For each generated word, we visualize the attention weights on individual pixels, outlining the region with the maximum attention weight in orange. Moreover, for each word, we display the corresponding visual weight of second residual based attention layer. From Fig.~\ref{fig.attention}, we find that our first layer attention is able to locate the right objects, which enables the DA+De+RF to accurately describe objects occurred in the input image. 
%Moreover, we also demonstrate the second layer visual attention weights, which are used for final answer prediction. From Fig.~\ref{fig.attention}, 
On the other hand, the visual weights in the second layer are obviously higher when our model predicts words related to objects (e.g., car and pizza).

\begin{figure}[t]
	\centering
	\scriptsize
	\includegraphics[width=0.95\linewidth]{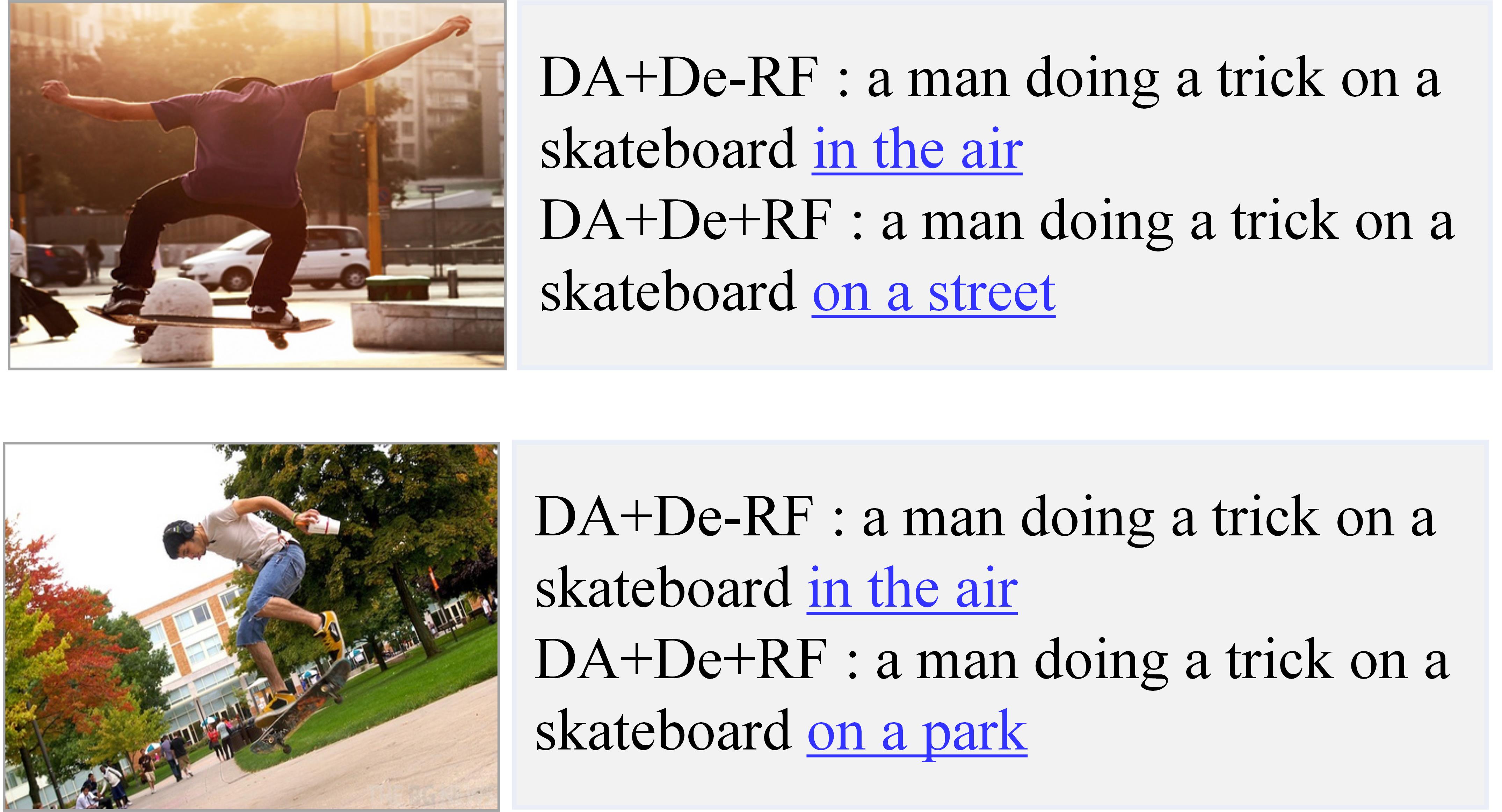}
	\caption{The role of RF. Examples of image captions generated by DA+De-RF and DA+De+RF.}
	\label{fig.disc}
\end{figure}

\noindent\textbf{The Influence of Reinforcement Learning.} From Tab.~\ref{tab.ablation}, we can see that the results clearly show the advantage of our reinforcement learning. From the first block of Tab.~\ref{tab.ablation}, we can see that DA-De+RF achieves 4.2\% BLUE1, 1.5\% BLUE4, 0.9\% METEOR, 1.5\% ROUGE, 12.4\% CIDEr and 1.7\% SPICE performance gain compared with DA-De-DF. Moreover, compared with DA+De-RF, DA+De+DF performs better with an increase of 4.1\% BLUE1, 1.8\% BLUE4, 1.1\% METEOR, 2.0\% ROUGE, 13.7\% CIDEr and 1.8\% SPICE. The increases of performance clearly demonstrate the effectiveness of the proposed reinforcement learning component. %In general, the improvement of DA+De+DF vs DA+De-DF is higher than the improvement of DA-De+DF vs DA-De-DF. This confirms the effectiveness of the combination between the deliberation procure with reinforcement learning. This combination is beneficial for image captioning.

\begin{table*}[t]
	\centering
	\caption{Results on the online MSCOCO test sever. DA is a single model, both SCST:Att2all and Up-Down are an ensemble of 4 models. LSTM-A3 utilizes Resnet-152 based visual feature while DA uses RestNet101 based visual feature.}
	\label{ONLINEe}
	\begin{tabular}{lllllllllllllll}
		\hline
		& \multicolumn{2}{c}{BLEU-1} & \multicolumn{2}{c}{BLEU-2} & \multicolumn{2}{c}{BLUE-3} & \multicolumn{2}{c}{BLUE-4} & \multicolumn{2}{c}{METEOR} & \multicolumn{2}{c}{ROUGE-L} & \multicolumn{2}{c}{CIDEr} \\ \cline{2-15}
		& \multicolumn{1}{c}{c5} & \multicolumn{1}{c}{c40} & \multicolumn{1}{c}{c5} & \multicolumn{1}{c}{c40} & \multicolumn{1}{c}{c5} & \multicolumn{1}{c}{c40} & \multicolumn{1}{c}{c5} & \multicolumn{1}{c}{c40} & \multicolumn{1}{c}{c5} & \multicolumn{1}{c}{c40} & \multicolumn{1}{c}{c5} & \multicolumn{1}{c}{c40} & \multicolumn{1}{c}{c5} & \multicolumn{1}{c}{c40} \\ \hline
		Review Net       & 72.0         & 90.0        & 55.0         & 81.2        & 41.4         & 70.5        & 31.1         & 59.7        & 25.6         & 34.7        & 53.5         & 68.6         & 96.5        & 96.9        \\
		Adaptive         & 74.8         & 92.0        & 58.4         & 84.5        & 44.4         & 74.4        & 33.6         & 63.7        & 26.4         & 35.9        & 55.5         & 70.5         & 104.2       & 105.9       \\
		PG-BCMR          & 75.4         & -           & 59.1         & -           & 44.5         & -           & 33.2         & -           & 25.7         & -           & 55           & -            & 101.3       & -           \\
		SCST:Att2all     & 78.1         & 93.7        & 61.9         & 86.0        & 47.0         & 75.9        & 35.2         & 64.5        & 27.0         & 35.5        & 56.3         & 70.7         & 114.7       & 116.7       \\
		LSTM-A3          & 78.7         & 93.7        & 62.7         & 86.7        & 47.6         & 76.5        & 35.6         & 65.2        & 27.0         & 35.4        & 56.4         & 70.5         & 116.0       & 118.0       \\ 
		Up-Down          & \textbf{80.2}         & \textbf{95.2}        & \textbf{64.1}         & \textbf{88.8}        & \textbf{49.1}        & \textbf{79.4}        & \textbf{36.9}         & \textbf{68.5}        & 27.6         & 36.7        & 57.1         & \textbf{72.4}         & 117.9       & \textbf{120.5}      \\
		DA & 79.4         & 94.4        & 63.5         & 88.0        & 48.7         & 78.4        & 36.8         & 67.4        & \textbf{28.2}         & \textbf{37.0}        & \textbf{57.7}         & 72.2         & \textbf{120.5}       & 122.0       \\ \hline
	\end{tabular}
\end{table*}

\begin{figure*}[t]
	\centering
	\scriptsize
	\includegraphics[width=1\linewidth]{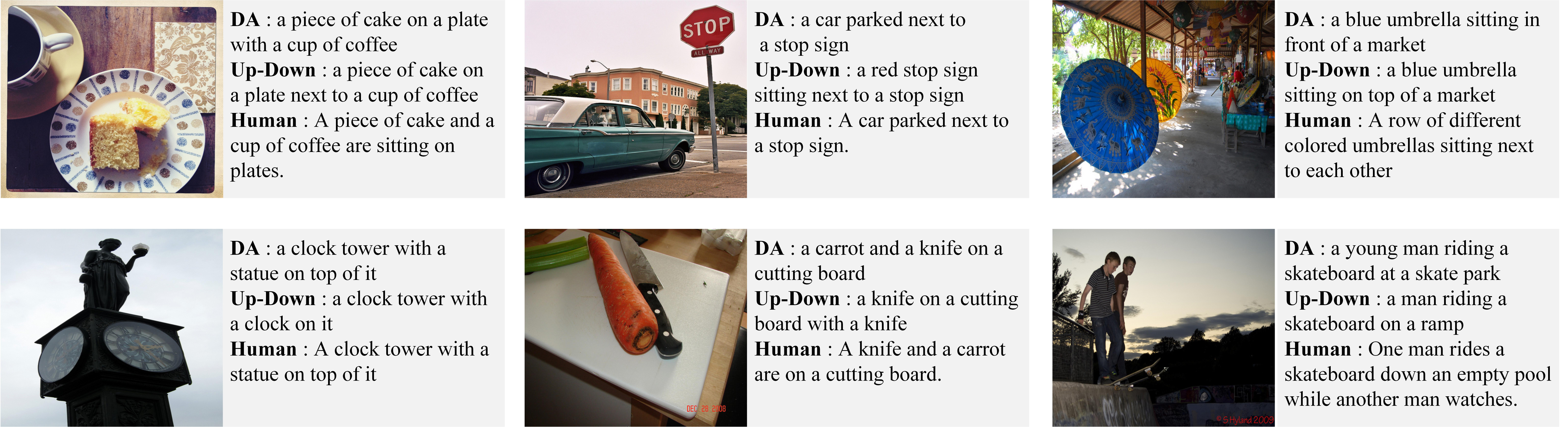}
	\caption{Examples of image captions generated by Up-Down, DA and human beings. The first column, both Up-Down and our DA provide accurate descriptions. The middle column shows that in some cases our DA is able to provide better descriptions, while the third column indicates that in complex situations both Up-Down and DA fail.}
	\label{fig.captions}
\end{figure*}

In order to further demonstrate the discriminability, we show some qualitative results in Fig.~\ref{fig.disc}. By observing the two images, a human can aware that two men are performing the same activity (i.e., doing a tricks on a skateboard), but they are playing at different places (i.e., on a street and on a park). The first man plays skateboard on the street, while the second man plays skateboard on a park. From Fig.~\ref{fig.disc}, we can see that DA+De-RF generates the same description for both images, while DA+De+RF is able to generate precise caption to describe their difference (i.e., on a street or on a park). This indicates the reinforcement learning with contrastive loss improves the discriminability of our DA model.

\subsection{Comparing with State-of-the-Art}
In this section, we use DA to represent our model DA+De+RF for convenience.

\textbf{Flickr30K.} For Flickr30K dataset, we compare our DA with DeepVS~\cite{deepvs}, Hard-Attention~\cite{hard_attention}, ATT-FCN~\cite{att_fcn} and Adaptive-Attention~\cite{adaptive} and the comparison results are shown in Tab.~\ref{tab.flickr}. From Tab.~\ref{tab.flickr}, we can see that our DA outperforms the counterparts by a large margin. Specifically, compared with Adaptive \cite{adaptive}, DA has 6.1\% BLUE-1, 5.7\% BLUE-2, 4.9\% BLUE-3, 4.3\% BLUE-4, 2.6\% METEOR and 13.5\% CIDEr increases. The improvement is significant, especially for BLUE-n and CIDEr.

\textbf{COCO.} We conduct two types of evaluations on the COCO dataset. The first is conducted offline by using the ``Karpathy'' split that have been widely used in prior work. The second one is conducted online and the captioning results are obtained on the online MSCOCO test server.

For offline evaluation, all the image captioning models are single-model. Here, we compare DA with SCST:Att2in~\cite{self_critical}, SCST:Att2all~\cite{self_critical}, ATTN+C+D(1)~\cite{discriminability} and Up-Down~\cite{bottomup}. The offline evaluation results are shown in Tab.~\ref{tab.coco}. It is clear that our DA performs the best on the widely used evaluation metrics, e.g., BLUE4, METEOR, ROUGE, CIDEr and SPICE scores. Up-Down~\cite{bottomup} is a strong competitor, and it performs the best for ``Card'' and ``Rel''.

For online evaluation, we compare with previous published works, including Review Net \cite{reviewNet}, Adaptive \cite{adaptive}, PG-BCMR \cite{PG-BCMR}, SCST:Att2all~\cite{self_critical}, LSTM-A3 \cite{LSTM-A3}, Up-Down~\cite{bottomup}. The experimental results are shown in Tab.~\ref{ONLINEe}. From Tab.~\ref{ONLINEe}, we can see that beside METEOR, Up-down in general performs the best. In fact, both Up-Down and SCST:Att2All are an ensemble of 4 models, while our DA uses a single-model. Although LSTM-A3 utilizes better visual features extracted from the ResNet-152, our DA with ResNet-101 visual features obtains higher performances, especially the METEOR scores reaching 28.2\% on c5 and 37.0\% on c40. We believe that the performance of DA could be further boosted via an ensemble of multiple DA-based models.

\subsection{Qualitative Analysis}
Here, we show some qualitative results in Fig.~\ref{fig.captions}. From the above Tables (i.e., Tab. \ref{tab.flickr}, Tab. \ref{tab.coco} and Tab.~\ref{ONLINEe}), we can see that the previously Adaptive \cite{adaptive} performs the best on the Flicker30k while Up-Down~\cite{bottomup} performs the best on the COCO dataset. Due to the reason that Up-Down~\cite{bottomup} releases the code while Adaptive \cite{adaptive} does not, we show some captioning examples obtained by our DA and Up-Down. The first column with two examples show that both DA and Up-Down are able to provide accurate description. For the middle column, we can see that our DA provides more accurate descriptions, especially for describing the relationships among objects. In this case, Up-Down fails to detect all objects and the relationships among them. The third column shows two images and both of them have complex background and their corresponding descriptions contain rich semantic information. For those two images, both  DA and Up-Down fails. One possible reason is that a human being can conduct reasoning based on his or her background knowledge while at this stage both DA and Up-Down cannot. This proposes a potential research direction for image captioning.

\section{Conclusion and Future Work}
\label{sec.conclusion}
In this paper, we firstly introduce a novel hLSTMat encoder-decoder framework, which integrates a hierarchical LSTMs, temporal attention and adaptive temporal attention to automatically decide when to make good use of visual information or when to utilize sentence context information, as well as to simultaneously considering both low-level video visual features and language context information. Experiments show that hLSTMat achieves state-of-the-art performances on both MSVD and MSR-VTT datasets. Secondly, we propose two spatial-temporal networks, namely ParA and Two-stream, and they further improve the performance of video captioning on all three datasets.

\ifCLASSOPTIONcompsoc
  \section*{Acknowledgments}
  
\else
  \section*{Acknowledgment}
\fi

This work is supported by the Fundamental Research Funds for the Central Universities (Grant No. ZYGX2014J063, No. ZYGX2014Z007) and the National Natural Science Foundation of China (Grant No. 61502080, No. 61632007, No. 61602049).

\ifCLASSOPTIONcaptionsoff
  \newpage
\fi

%\bibliographystyle{IEEEtran}


%\newpage


%\vfill


\end{document}